\documentclass[preprint,12pt]{elsarticle}

\usepackage{amssymb}
\usepackage{amsmath}
\usepackage{algorithm}
\usepackage[noend]{algpseudocode}
\usepackage{hyperref}
\usepackage{bm}
\usepackage{easyReview}

\journal{Journal of Computational Physics}

\begin{document}

\begin{frontmatter}



\title{A Surrogate-Augmented Symbolic CFD-Driven Training Framework for Accelerating Multi-objective Physical Model Development}


\author[melb]{Yuan Fang\corref{cor1}\fnref{equal}}
\author[melb]{Fabian Waschkowski\fnref{equal}}
\author[melb]{Maximilian Reissmann}
\author[melb]{Richard D. Sandberg}
\author[japan]{Takuo Oda}
\author[japan]{Koichi Tanimoto}
\cortext[cor1]{Corresponding author:fang.y5@unimelb.edu.au}
\fntext[equal]{These authors contributed equally to this work.}

\affiliation[melb]{organization={Department of Mechanical Engineering, University of Melbourne},%Department and Organization
            addressline={Parkville}, 
            city={Melbourne},
            postcode={3010}, 
            state={VIC},
            country={Australia}}
\affiliation[japan]{organization={Research and Innovation Centre Takasago Area, Mitsubishi Heavy Industries, Ltd,.},%Department and Organization
            addressline={2 Chome-1 Araicho Shinhama}, 
            city={Takasago},
            postcode={676-0008}, 
            state={Hyogo},
            country={Japan}}
%% Abstract
\begin{abstract}
% Background. 
% CFD-driven training is promising and how it works.
Computational Fluid Dynamics (CFD)-driven training, which combines machine learning (ML) techniques with fluid dynamics software that provides CFD feedback, has shown great potential for developing implementable and physically consistent closure models with improved accuracy compared to conventional baseline models. 
In the original CFD-driven training framework, each ML-generated candidate model is embedded in a CFD solver, and its performance is evaluated by comparing computed quantities of interest against reference data.
% Research Problem -> high computation cost
However, exploring optimal models requires hundreds to thousands of CFD evaluations, resulting in prohibitive training costs for complex industrial flows, where a single CFD calculation can already be computationally expensive. 
% Methodology -> couple surrogate modelling in symbolic CFD-driven training framework
To address this limitation, we propose an extended framework that, for the first time, integrates surrogate modeling into symbolic CFD-driven training in real time, aiming to reduce the overall training cost. The surrogate model learns to approximate the errors of the ML-generated models based on previous CFD evaluations, and this surrogate mapping is continuously refined throughout the training process. Newly generated models are first assessed by the surrogate. Only those predicted to yield small errors or high uncertainty are subsequently re-evaluated with full CFD computations. Specifically, the generated discrete expressions from symbolic regression are mapped into a continuous space using the averaged values of the input symbols as inputs to the probabilistic surrogate model. To accommodate multi-objective model training, particularly in situations where assigning fixed weights to competing quantities is challenging, the surrogate model is further generalized to multi-output settings. This is achieved by extending the kernel function to a matrix form, enabling one mean and variance prediction per training objective. Based on these probabilistic outputs, various selection metrics and thresholds are evaluated to identify the optimal setup.
% Results -> flow cases introduction and increased training efficiency
The proposed surrogate-augmented CFD-driven training framework is developed and demonstrated across a range of cases, encompassing statistically one- and two-dimensional flows, as well as single-expression (targeting only the turbulence model) and multi-expression (simultaneous optimization of both turbulence and heat-flux models). Across all cases, the framework achieves a substantial reduction in training cost while the resulting models maintain comparable predictive accuracy to those obtained with the original CFD-driven approach.
\end{abstract}

 \begin{graphicalabstract}
\includegraphics[width=1\linewidth]{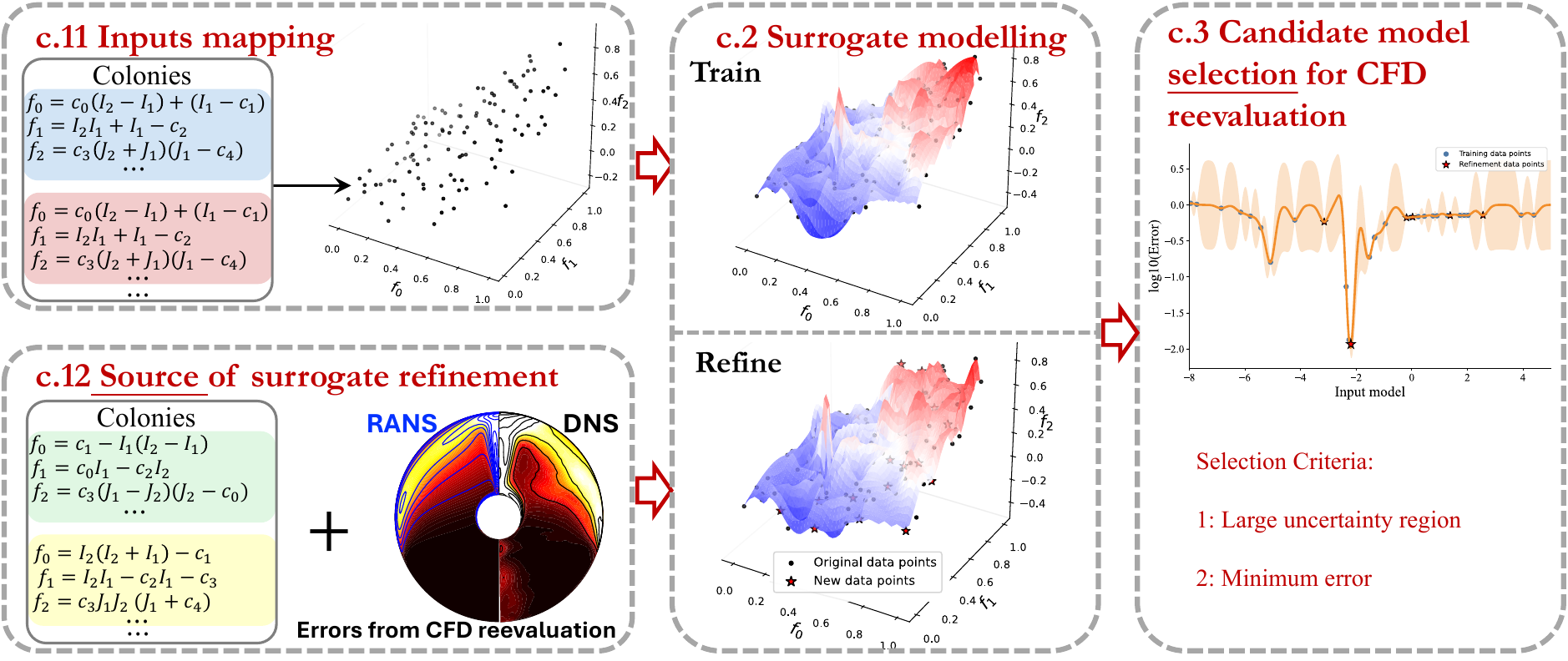}
\end{graphicalabstract}

\begin{highlights}
\item Integrated surrogate modeling into a symbolic-regression CFD-driven training framework for physical-model development
\item Verified the framework’s robustness across varied hyperparameter settings and flow cases involving turbulence alone or coupled with heat flux.
\end{highlights}

%% Keywords
\begin{keyword}
Surrogate modeling \sep Machine learning \sep Turbulence modeling \sep Heat flux modeling \sep Gaussian Processes


\end{keyword}

\end{frontmatter}



\section{Introduction}
\label{sec:introduction}
% Background
% RANS is still needed -> the accuracy relys on closure terms -> the success of CFD-driven training
%Reynolds-Averaged Navier-Stokes (RANS) calculations remains the workhorse of both industry and academia owing to its substantial computation cost savings compared with high-fidelity simulations such as large eddy simulations (LES) and direct numerical simulations (LES), as highlighted in the computational fluid dynamics (CFD) vision 2030 road map~\cite{cary2021cfd}. The efficiency of RANS stems from solving only the governing equations of the mean flow together with a limited set of turbulent quantities - such as turbulent kinetic energy, specific dissipation rate, turbulent dissipation rate, turbulent viscosity etc.. The influence of turbulence on mean-flow behaviour is then modelled through closure terms expressed as functions of these mean-flow and turbulent quantities. In contrast, LES and DNS explicitly resolve large or all turbulent scales with small time stepping, respectively, at far higher computational expense. Consequently, the predictive accuracy of RANS critically depends on the assumptions underlying its closure terms. 

% Background: CFD-driven training, cite zhao and fabian's work has gain great success, introduce the algorithm advantage and application. then, the huge training cost constraints the application on more complex problems, multi-expression, multi-objective, multi-case training, so on. Meanwhile, the models information generated during the training ls very limited.
Among the various efforts to incorporate machine learning (ML) into physical model development, Computational Fluid Dynamics (CFD)-driven training has demonstrated remarkable and continuing success \cite{zhao2020rans, saidi2022cfd,waschkowski2022multi}. In this approach, ML algorithms are directly coupled with CFD solvers, so that each ML-generated candidate model is embedded in the solver and evaluated via full CFD calculations. The CFD outputs are then used to compute the cost function, which serves as feedback to the ML algorithm. The training process continues iteratively until the cost function converges to a predefined threshold or the maximum number of iterations is reached. This tight integration offers two advantages: Firstly, the trained models exhibit strong robustness and physical consistency, since they are validated within the CFD solver rather than merely fitted to post-processed training goals \cite{wu2019reynolds}. Secondly, the cost function can be flexibly defined in terms of quantities of engineering interest beyond the closure terms themselves, allowing the model to be trained to target any physically meaningful flow quantities that can be extracted from the CFD solution. 
These advantages enable the development of turbulence~\cite{zhao2020rans,saidi2022cfd}, heat flux~\cite{xu2021data}, transition~\cite{fang2024data} models in Reynolds-Averaged Navier-Stokes (RANS) calculations, or LES sub-grid-scale~\cite{reissmann2021application} and wall~\cite{Cato2025} models using only limited reference data, thus even sparse experimental data can be employed for model training. 

% Research Problem
% Current training limits on simple flows due to the high training cost
Extensions to the CFD-driven training framework have enabled the development of physical models for increasingly complex flow scenarios, but have also increased training costs. Waschkowski~\cite{waschkowski2022multi} developed two such functionalities, named multi-expression and multi-objective training. The former allows multiple models that interactively affect the final flow prediction, such as coupled transition and turbulence modeling~\cite{fang2024data} or joint heat flux and turbulence modeling~\cite{xu2022towards}, to be trained simultaneously. The latter enables consideration of multiple quantities of interest in constructing the cost function without predefining their relative weights, thereby avoiding bias in the optimization process~\cite{fang2024data,harshal2022}. Moreover, to improve the generalizability of ML-trained models, multi-case training has been proposed to train models across multiple flow cases with varying geometries and operation conditions~\cite{fang2023toward}. Although these extensions significantly expand the applicability of CFD-driven training to complex and realistic problems, they also enlarge the search space, thereby increasing training costs. For example, if the ML algorithm initially generates 100 candidate models and subsequently updates 50 models in each of the following 100 training iterations, assuming that one CFD evaluation requires one core hour, the total computational cost amounts to $(100+50\times99)\times1=5,050$ core hours. This severely limits the applicability of the CFD-driven training framework to more complex, industry-related cases, such as unsteady, three-dimensional, or multiphase flows, where even a single CFD calculation is computationally expensive. Therefore, the extremely high training cost remains a significant barrier to the broader application of CFD-driven training.

% why surrogate model suits our problem with two reasons, unused generated models' information and the surrogate model algorithm itself 
To mitigate this prohibitive computational cost, this study introduces surrogate modeling into the CFD-driven training framework.  Surrogate modeling techniques have demonstrated great potential in design optimization problems \cite{li2022machine, schouler2025comparison, fruzza2026mapping}, uncertainty quantification \cite{sudret2017surrogate,tripathy2018deep} and inverse modeling \cite{asher2015review,mirghani2012enhanced}, where they effectively bridge the gap between limited high-fidelity data and low-fidelity predictions. For CFD-driven training, the surrogate model is designed to exploit the wealth of information contained in previously generated models, including those with relatively large errors that are typically discarded, and to leverage this information to reduce training cost. 
It is expected that as the training progresses, many candidate models become fine-tuned and share similar structures and comparable performance, making full CFD evaluations for all of them unnecessary and computationally expensive. Surrogate modeling, which constructs a computationally efficient approximation of a complex and expensive system using data from a limited number of prior evaluations, therefore naturally fits this problem. By learning from accumulated CFD results from earlier training and continuously refining itself during subsequent iterations, the surrogate model can predict the performance of most newly generated candidates, allowing full CFD evaluations to be reserved for those with high potential or considerable uncertainty. This integration is expected to reduce overall training costs while maintaining substantial predictive accuracy.

% literature review on surrogate model methodology to introduce the Gaussian processes used in our study
Various surrogate modeling approaches have been proposed in the literature. 
Traditional polynomial response surface models approximate the system response using low-order polynomials in the input variables; however, they suffer from the curse of dimensionality, as the number of polynomial terms grows with the number of inputs. This often leads to overfitting and limits their capability to capture complex nonlinear relationships \cite{bilionis2013multi,atkinson2019structured}. Radial basis function (RBF) models, on the other hand, offer good interpolation properties through localized basis functions, but are highly sensitive to kernel parameters \cite{majdisova2017radial}. Other machine-learning-based surrogates, such as artificial neural networks (ANNs)~\cite{sun2019review}, support vector regression (SVR)~\cite{shi2020multi}, and random forests~\cite{dasari2019random}, can learn complex nonlinear and high-dimensional mappings but generally lack interpretability.

In contrast, Gaussian Process (GP) modeling provides a probabilistic and interpretable framework for surrogate construction and is adopted in this study because of its simplicity, flexibility, and ability to quantify uncertainty~\cite{kennedy2000predicting}. A GP can be viewed as a generalization of a Gaussian probability distribution to functions, where any finite collection of function values follows a multivariate Gaussian distribution characterized by a mean vector and a covariance matrix. The predictive mean gives the estimated output (e.g., the cost function value), while the predictive variance quantifies the associated uncertainty and naturally reflects confidence in each prediction. This property makes GPs particularly suitable for the present problem, where the surrogate model predicts both the cost function value and its confidence interval for each candidate model. Consequently, only models predicted to have very small errors or large uncertainty are re-evaluated using full CFD simulations, thereby significantly accelerating the overall training process.

% Contribution
In this study, the GP-based surrogate modeling technique is integrated into our existing CFD-driven training framework~\cite{waschkowski2022multi}. 
% To the best of our knowledge, although surrogate modeling has been previously employed as a mapping function between high- and low-fidelity predictions ~\cite{schouler2025bayesian,fairbanks2020bi,lee2025surrogate,   lu2024application}, it has not yet been coupled with a CFD-driven training framework to accelerate the model development process. 
The proposed framework, referred to as the surrogate-augmented CFD-driven training framework, extends the original CFD-driven training framework, and these two terms will be used throughout this paper. After the first training generation, the candidate models and their corresponding errors obtained from CFD evaluations are used to train the surrogate model. In subsequent generations, only the candidate models predicted by the surrogate to have very small errors or considerable uncertainty are re-evaluated through full CFD simulations. The updated results from these evaluations are then incorporated as additional data to refine the surrogate model. In this manner, the surrogate model evolves concurrently with the CFD-driven training, continuously improving its predictive capability. Further methodological details are provided in Section {sec:methodology}.

% Paper structure
This paper is organized as follows. Section 2 describes the establishment of the surrogate modeling framework for physical model development, its coupling with the existing symbolic-regression-based CFD-driven training framework, and the extension to multi-objective surrogate-based training. Section 3 introduces the physical model closures and flow cases employed in this study. Section 4 presents the comparison of training efficiency between the regular and surrogate-augmented CFD-driven training frameworks, along with the explicit model expressions and performance evaluations. Finally, Section 5 provides the conclusions and outlines future perspectives.

\section{Methodology}
The methodology section is organized into two main parts. The first part describes the integration of surrogate modeling into the current CFD-driven training framework, detailing each component, including the input parameters, the surrogate modeling algorithm, and the corresponding outputs. The second part introduces the physical closure models derived from ML training, along with descriptions of all the test cases used in this study.

\subsection{Surrogate Modeling Implementation}
\begin{figure}[t]
\centering
\includegraphics[width=1\linewidth]{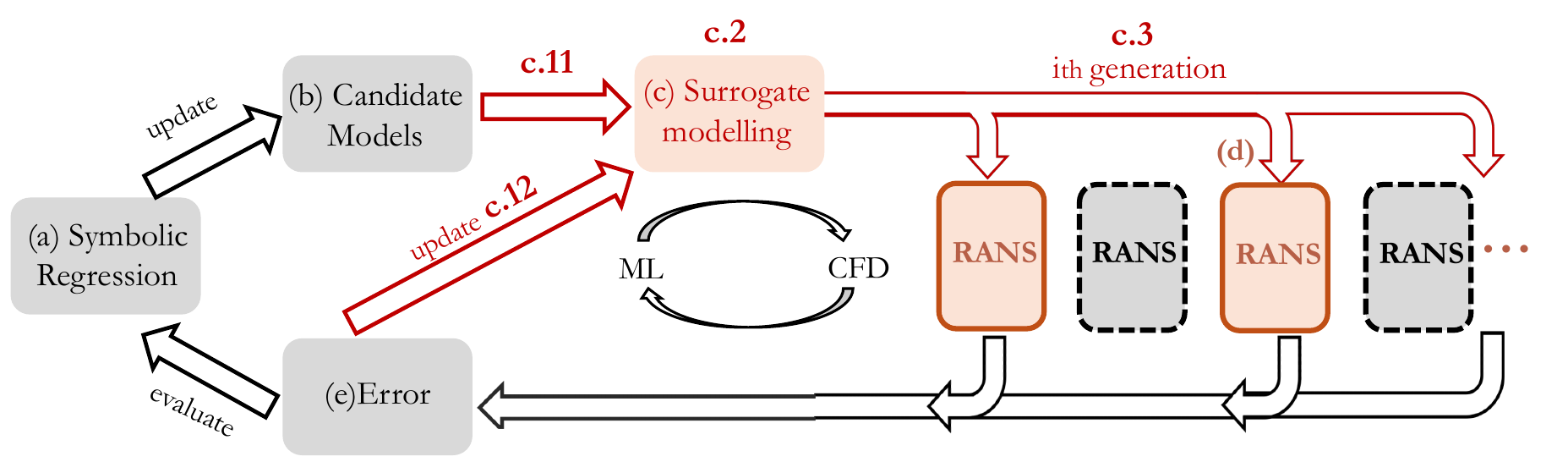}
\caption{
Surroate-augmented CFD-driven training framework Newly introduced components relative to the previous framework are highlighted in red, while elements retained from the original framework are shown in grey. Grey boxes outlined with black dashed lines indicate RANS simulations that are skipped (not executed) and replaced by surrogate predictions.}\label{fig:framework}
\end{figure}
Figure~\ref{fig:framework} illustrates the surrogate-augmented CFD-driven training framework, where the components of the regular CFD-driven framework~\cite{zhao2020rans,waschkowski2022multi} are shown in black, and the newly developed surrogate modeling components are highlighted in red.
The model training process begins with regular CFD-driven training, followed by the introduction of the surrogate modeling algorithm in subsequent training iterations. This design is necessary because, in the initial iteration, no candidate models nor corresponding error data are yet available for training the surrogate model. For each generation, the symbolic regression ML method (in this study, GEP) generates candidate models, as shown in Figures~\ref{fig:framework} (a). These models in Figure~\ref{fig:framework} (b) are then used as inputs to the surrogate model in Figure~\ref{fig:framework} (c), which predicts both the expected error and the associated uncertainty range for each model. Only those models predicted to yield minor errors or exhibit large uncertainty are selected for CFD re-evaluation, as illustrated in Figure~\ref{fig:framework} (d). The re-evaluated models, together with their actual CFD errors, are subsequently used to refine the surrogate model.
Meanwhile, the combined information from the surrogate predictions and CFD evaluations is fed back to GEP in Figure~\ref{fig:framework} (a) to guide further model evolution. This iterative training process continues until either the model error falls below a prescribed threshold or the maximum number of training iterations is reached. In the following subsections, each component relevant to the surrogate modeling development is discussed in detail.

\begin{figure}[t]
\centering
\includegraphics[width=1\linewidth]{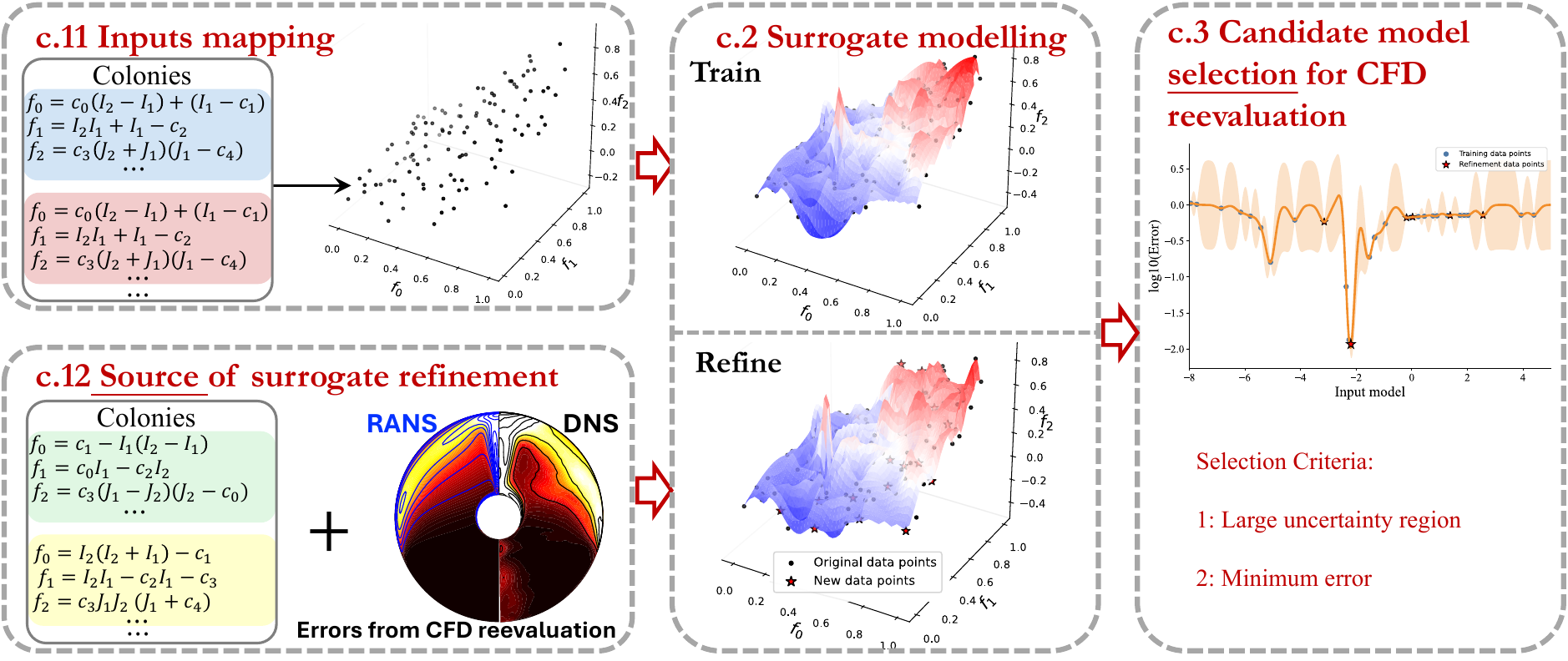}
\caption{Main components of the surrogate model framework integrated into CFD-driven closure training}\label{fig:surrogateComponents}
\end{figure}
\subsubsection{Inputs mapping}
% first explain the problem and then solution
The input to the surrogate modeling is the output from the GEP approach — the candidate models that are represented as distinct character strings, as shown in Figure~\ref{fig:framework}. These strings possess no inherent notion of smooth change between each other; in other words, they are discrete in nature. However, the surrogate model used in this study is based on Gaussian processes, which typically operate on continuous spaces, where closeness in the input space implies correlation in the output. Therefore, the discrete inputs generated by the GEP must first be mapped into a continuous representation, as illustrated in Figure~\ref{fig:surrogateComponents} (c.11), before being fed into the surrogate modeling algorithm. The mapping is required to represent the similarity between the generated models through their relative distances in a continuous space. 

We now describe the procedure for transforming a discrete symbolic expression into a continuous representation, followed by a description of how these high-dimensional embeddings are projected onto a lower-dimensional manifold to extract semantically relevant information. The detailed implementation of this process is presented in Algorithm~\ref{alg:mapping}.

\begin{algorithm}[H]
\caption{Mapping symbolic phenotypes to a continuous low-dimensional representation}
\label{alg:mapping}
\begin{algorithmic}[1]
\Require Phenotype expressions $\{f_k\}^n_{k=1}$, input values (from base line model calculation)$\{\mathbf{I},\mathbf{J}\}_{i=1}^m$
\Ensure Mapped phenotype expressions $\{h(f_k)\}_{k=1}^{n}$
\For{$f \in \{f_k\}_{k=1}^n$ }
     \For{$I,J\in \{\mathbf{I},\mathbf{J}\}_{i=1}^m$ }
        \State $y_i=f(I,J)$  \Comment{Point-wise model evaluation}
     \EndFor{\textbf{end for}}
     \State $y = \frac{1}{m}\sum_iy_i$ \Comment{Dimensional reduction via aggregation}
     \State $\{h(f)\}_i \gets y$ \Comment{Append to return value}
\EndFor{\textbf{end for}}
\end{algorithmic}
\end{algorithm}
With each model population, a list of expressions, each expression (simplified phenotype representation) is evaluated pointwise on a dataset (steps 1-3). Specifically, the GEP framework constructs a model equation by assembling symbols from a predefined set, e.g., $\mathcal{S} = {I_i, I_2, +, -, \times}$ into a list referred to as the genotype. For example: 
\begin{equation}
\label{eqn:geno}
\times \: + \: I_1 \: I_1 \: I_2
\end{equation}
represents a genotype that the GEP algorithm decodes into an expression tree (via depth-first pre-order traversal), which can be interpreted as the corresponding phenotype equation, i.e. $(I_1+I_1)\times I_2$ for Equation~\ref{eqn:geno}. This expression is utilized because altering a single symbol in the genotype can substantially alter the resulting phenotype—and, consequently, the CFD simulation results and model error. For instance, changing Equation \ref{eqn:geno} to $\times \: \times \: I_1 \: I_1 \: I_2$ modifies the phenotype from $2I_1\times I_2$ to $I_1^2\times I_2$. Conversely, another genotype such as $\times \: I_2 \: + \: I_1 \: I_1$ differs syntactically from Equation \ref{eqn:geno} but encodes the same phenotype. Hence, phenotype-based strings are selected for subsequent processing and simplified to remove redundancy.

Up to this point, the phenotype-based input strings remain high-dimensional. Each symbol is represented using one-hot encoding, where a symbol set of size $N$ is expressed as an $N$-dimensional vector with a value of 1 at the position corresponding to the active symbol and zero elsewhere. Steps 2–4 describe how this information is projected into a lower-dimensional, more learnable space for the surrogate model. In Step 2, the values of input symbols, such as $I_i$, are extracted from baseline model calculations, as high-fidelity experimental or simulation datasets are often unavailable for complex, industry-relevant flows. In Step 3, the corresponding values of the generated models at the same locations are computed. In Step 4, an aggregation function $h(y)$ is applied to reduce the model predictions to a low-dimensional representation; here, the mean model prediction is used. Consequently, each generated model corresponds to a single point in the continuous space, as illustrated in Figure \ref{fig:surrogateComponents} (c.11).

\subsubsection{Algorithm and the hyperparameters}
The task of the surrogate modeling algorithm is, given the mapped input values of a candidate model, to predict its corresponding error and quantify the uncertainty of this prediction, as illustrated in Figure~\ref{fig:surrogateComponents} (c.2). Gaussian Processes (GPs) provide these quantities through the mean $\mu$ and variance $\sigma^2$. Specifically, a GP represents realizations of a random function as a vector of random variables following a multivariate Gaussian distribution characterized by a mean vector $\boldsymbol{\mu}$ and a covariance matrix $\boldsymbol{\Sigma}$. Specifying these two quantities defines the GP and establishes a prior distribution over functions. 

For the mean function $\mu(x)$, a zero-mean assumption is commonly adopted, implying that the expected value of the random function over all sample functions at a specific point is zero. Consequently, the covariance function $\Sigma(x, x’)$, also referred to as the kernel of the GP, governs the correlations between function values and thereby defines the prior distribution over functions. By combining this prior with the available training data through Bayes’ theorem, a posterior distribution is obtained that captures both the data trends and predictive uncertainty. The essential step in constructing a GP model, therefore, is to specify an appropriate kernel function.

Once the kernel is defined, no explicit parameter learning is required, which is why GPs are regarded as nonparametric models. The kernel function examined in this work is the Rational Quadratic (RQ) kernel:
%In this study, the kernel functions are investigated to assess their influence on the surrogate model performance. One of the kernel functions employed in this study is the Radial Basis Function (RBF) kernel, which is the most widely used and often considered the default GP kernel in the literature:

%\begin{equation}
%\label{eqn:RBF}
%\Sigma (x,x^{'}) = \sigma^{2} e^{\left (-\frac{\lVert x-x^{'}\rVert^2}{2\ell^2} \right)}
%\end{equation}
%where $x, x^{'}$ denote two input data points, with the two hyperparameter $\theta = {\sigma, \ell}$. Here, $\sigma$ is the vertical amplitude to scale output variance, while $\ell$ defines the length scale, determining how rapidly the correlation decays with increasing distance between $x$ and $x^{'}$. 
\begin{equation}
\label{eqn:RQ}
\Sigma (x,x^{'}) = \sigma^{2} \left (1+\frac{\lVert x-x^{'}\rVert^2}{2\alpha \ell^2} \right)^{-\alpha}
\end{equation}
which introduces an additional shape parameter $\alpha$, resulting in the hyperparameter set $\theta = {\sigma, \ell, \alpha}$.
The parameter $\alpha$ controls the extent to which the effective length scale $\ell$ varies, allowing the RQ kernel to model functions with different degrees of smoothness across the input space.

The above hyperparameters are determined by maximizing the logarithmic marginal likelihood of the observed data, $p\left ( y|X,\theta \right)$, defined as
\begin{equation}
\label{eqn:log}
\theta^* = \mathrm{argmax}_{\theta} \mathrm{log}\:p\left ( y|X,\theta \right)
\end{equation}
where $X$ denotes the input data points and $y$ represents the corresponding model outputs.

To enable multi-objective optimization within the surrogate-augmented CFD-driven training framework, the surrogate model is extended to predict multiple output quantities. Specifically, for each training objective, the model provides one mean $\mu$ and one variance prediction $\sigma^2$. For a total of $p$ outputs, the GP kernel is generalized to a matrix form as follows:
\begin{equation}
\label{eqn:multiKernel}
\boldsymbol{\Sigma}(x, x') =
\begin{bmatrix}
\Sigma_{11}(x, x') & \cdots & \Sigma_{1p}(x, x') \\
\vdots & \ddots & \vdots \\
\Sigma_{p1}(x, x') & \cdots & \Sigma_{pp}(x, x')
\end{bmatrix}
\end{equation}
where the diagonal element $\Sigma_{ii}(x, x^{'})$ represents the covariance function associated with the $i^{th}$ training objective, where the off-diagonal elements $\Sigma_{ij}(x, x^{'})\: (i \neq j)$
 describe the cross-covariances between different output features. Inference with multi-output GPs requires inverting the full covariance matrix that incorporates all $n$ training samples, resulting in a matrix of size $np \times np$ for $p$ output features. Since matrix inversion scales cubically in time, $\mathcal{O}(n^{3}p^{3})$, and quadratically in memory, $\mathcal{O}(n^{2}p^{2})$, the computational cost can quickly become prohibitive. In the present study, we assume independence among the training objectives and set $\Sigma_{ij}(x, x^{'}) = 0~for~i \neq j$.

\subsubsection{Candidate Model Selection for CFD reevaluation}
\label{subsection:model selection}
The predicted mean $\mu$ of the surrogate model approximates the true error values of the candidate models and replaces direct CFD evaluation. The variance $\sigma^2$ from the surrogate model, on the other hand, represents the predictive uncertainty, and can be utilized in combination with the error estimates to identify the candidate models that require re-evaluation through CFD calculation, as shown in Figure~\ref{fig:surrogateComponents} (c.3). The goal is to achieve the same minimum fitness values as in regular CFD-driven training while performing as few CFD evaluations as possible. 

\begin{algorithm}
\caption{Selection of candidate models for CFD re-evaluation}
\label{alg:selection}
\begin{algorithmic}[1]
\Require Training generation index $n$, model population $P$, and hyperparameters $\mathcal{H}$
\If{$n = 0$} 
    \Comment{No CFD data available to train the surrogate model}
    \State $\mathbf{P}_{0} \gets \text{errors from CFD predictions}$ 
    \Comment{Randomly initialize the starting population}
\ElsIf{$n = 1$} 
    \State $\mathbf{P}_{\text{sampled}} \gets \text{sample}(m)$ 
    \Comment{Resample $m$ individuals uniformly from the population}
\Else 
    \State $\mathbf{P}_{\text{sampled}} \gets \text{selection metric}$ 
    \Comment{Select based on predicted mean errors and uncertainty levels}
\EndIf
\State $\mathbf{P}_{\text{sampled}} \gets \text{CFD evaluations}$ 
\State $\mathbf{P} \setminus \mathbf{P}_{\text{sampled}} \gets \text{surrogate model predictions}$
\Comment{Only the selected models are re-evaluated using CFD}
\end{algorithmic}
\end{algorithm}

Initially, a random population of models is generated by the GEP algorithm. Since no CFD-based error data are yet available to train the surrogate model, all candidate models are evaluated using CFD simulations. In subsequent training generations, once error information from the initial population becomes available, a uniform distribution of $n_i$ sampling points is generated. To ensure balanced sampling when the surrogate model has limited training data and low prediction accuracy, a nearest-neighbor search is performed to identify the model closest to each sampling point. A fixed number of $n_i$ models is then selected for CFD re-evaluation, while the errors of the remaining models are predicted using the surrogate model. For all subsequent training generations, only candidate models that meet a predefined selection threshold based on a selection metric that combines surrogate-predicted error and uncertainty are re-evaluated by CFD. The remaining models are replaced by surrogate predictions, reducing computational cost while maintaining accuracy.

The key question that follows is: what selection metric and threshold should be used to determine which models undergo CFD re-evaluation? 

Three selection metrics were investigated and implemented, each assigning a selection value to each candidate model based on the surrogate model's mean and variance predictions. The first metric considered is the Lower Confidence Bound (LCB)
\begin{equation}
\label{eqn:LCB}
m_{LCB}(x) = -\mu(x) + \beta \sigma(x),
\end{equation}
which is defined as a linear combination of the predictive mean $\mu$ and standard deviation $\sigma$, weighted by a hyperparameter $\beta$. A value of $\beta = 0$ corresponds to selecting candidates with the smallest predicted errors, whereas a larger $\beta$ favors the selection of candidates that contribute most to reducing the predictive uncertainty.

The second metric considered is Expected Improvement (EI), one of the most widely used selection criteria in the literature. The EI metric quantifies the expected improvement of a candidate model over the current best-performing solution, balancing exploitation of promising regions and exploration of uncertain areas predicted by the surrogate model.
\begin{equation}
\label{eqn:I}
I(x) = \text{max}(0, f(x^+)-\mu(x)),
\end{equation}
where $f(x^+)$ denotes the error of the current best-performing candidate model in the population. The expected improvement can then be calculated as
\begin{equation}
\label{eqn:EI}
m_{EI}(x) = (f(x^+)-\mu(x)-\xi) \Phi\left(\frac{f(x^+)-\mu(x)-\xi}{\sigma(x)} \right) + \sigma(x) \phi\left(\frac{f(x^+)-\mu(x)-\xi}{\sigma(x)} \right),
\end{equation}
where $\Phi$ and $\phi$ represent the cumulative distribution function and the probability density function of a standard Gaussian distribution, respectively. The hyperparameter $\xi$ is introduced to artificially reduce the improvement $I(x)$, thereby encouraging exploration of regions with higher uncertainty.

When assigning selection values based on the selection metric, an additional treatment is applied to exclude non-convergent models from being selected. In the regular CFD-driven training, divergent models are typically marked by assigning a large error value (e.g., 9999) to ensure they are excluded from subsequent optimization rounds. In the present framework, a convergence weighting, denoted as $m_{\mathrm{CW}}$, is introduced to scale the selection values:
\begin{equation}
\label{eqn:CW}
m_{CW} = \text{min}\left (1, \frac{\lVert x-x_{\mathrm{div}}\rVert}{\delta\lVert x_{\mathrm{conv}}-x_{\mathrm{div}}\rVert} \right),
\end{equation}
where the convergence factor $\delta$ controls the onset of linear discounting. For example, when $\delta = 0.5$, the convergence weighting reaches unity for selection distances greater than $0.5 \lVert x_{\mathrm{conv}} - x_{\mathrm{div}} \rVert$, while models located closer to the divergence boundary receive smaller weighting values. 

For multi-objective optimization, the selection value is computed separately for each output feature of every candidate model, following a procedure similar to that proposed by Jeong and Obayashi \cite{jeong2005efficient}. These individual selection values are then integrated into a single scalar value per candidate using multi-objective optimization techniques.

After assigning selection values based on the error and uncertainty predicted by the surrogate model, selection thresholds must be defined to determine which models are chosen for re-evaluation. In total, four thresholding strategies are implemented:
\begin{enumerate}
    \item Fixed-number selection (\(m_{s,f}\)) --- selects a predefined number of models.
%    \item Absolute-value selection (\(m_{s,a}\)) --- selects models whose selection values exceed a preset absolute threshold.
    \item Relative-value selection (\(m_{s,r}\)) --- selects models whose normalized selection values exceed a specified relative threshold across the entire input space.
    \item Pareto-front selection (\(m_{s,p}\)) --- used in multi-objective optimization to select only the models located on the assigned Pareto front.
\end{enumerate}
These thresholds can also be applied in combination. For example, specifying \(m_{s,f} = 10\) and \(m_{s,r} = 0.5\) means that only ten models with normalized selection values greater than \(0.5\) are selected for CFD re-evaluation.

\subsection{Modeling strategy and description of flow cases}
This subsection first introduces the training objectives for the physical model development, together with the corresponding inputs and outputs. The developed framework is tested on both standalone turbulence-model training and coupled turbulence and heat-flux model training. Subsequently, the computational setup of the development and demonstration cases and the formulation of their associated cost functions are presented. The selected cases cover statistically one- and two-dimensional flows under a range of operating conditions.  

\subsubsection{Turbulence and heat flux modeling}
The Boussinesq hypothesis used in the two-equation turbulence model (here, $k-\omega$ SST) assumes that the Reynolds stress tensor $\bm{\tau}_{ij}$ is linear proportional to the deviatoric component of the mean strain rate $\bm{S}^*_{ij}$, an approximation that oversimplifies the underlying physics in highly anisotropic and non-equilibrium flow regions. The objective here therefore, is to extend the linear model with the supplement proposed by Pope~\cite{pope1975more}, i.e., with an extra anisotropy as:
\begin{equation}
\begin{aligned}
&\bm{\tau}_{ij} =  \underbrace{\frac{2}{3}\rho k \bm{\delta_{ij}}}_{\textrm{isotropy}}  -  \underbrace{2\mu_{T}\bm{S}^*_{ij}}_{\textrm{anisotropy}} + \underbrace{2 \rho k \bm{a}_{ij}}_{\textrm{extra anisotropy}}, \\
&\bm{a}_{ij}=g_1\bm{V}^1_{ij} + g_2\bm{V}^2_{ij} + g_3\bm{V}^3_{ij}, \\
&g_{i=1,2,3}=f(A[1:5], I_1, I_2, +, -, \times)
\end{aligned}
\label{eqn:tauijML}
\end{equation}

Three tensor bases $\bm{V}^1_{ij}=\bm{s}_{ij}$, $\bm{V}^2_{ij}=\bm{s}_{ik}\bm{w}_{kj}-\bm{w}_{ik}\bm{s}_{kj}$ and $\bm{V}^3_{ij}=\bm{s}_{ik}\bm{s}_{kj}-\frac{1}{3}\bm{\delta}_{ij}\bm{s}_{mn}\bm{s}_{nm}$ are adopted for statistically two-dimensional flows \cite{pope1975more, gatski1993explicit}. $I_1=\bm{s}_{mn}\bm{s}_{nm}; I_2=\bm{w}_{mn}\bm{w}_{nm}$ are two associated non-zero scalar invariants. In these definitions, $\bm{s}_{ij}=(1/\omega)\bm{S}^{'}_{ij}$ and $\bm{w}_{ij}=(1/\omega)\bm{\Omega}_{ij}$ denote non-dimensionalized strain and rotation rate tensors, respectively. The inputs for the turbulence model closure development include the constants $A$, two invariants $I_1, I_2$ and a set of mathematical operators, while the outputs are the coefficients $g_{i=1,2,3}$. 

An artificial turbulent production term, $R$, as introduced by \cite{schmelzer2020discovery}, is trained alongside the additional anisotropic stress. This term adopts a structure similar to the original turbulent production term, based on the observation that the anisotropic stress correction alone is insufficient to achieve the desired model accuracy.
\begin{equation}
\begin{aligned}
 R = 2kb_{ij}\partial_j U_i,
\end{aligned}
\label{eqn:extra}
\end{equation}
where $b_{ij}(V_{ij}^k, I_k)=g_4 V_{ij}^1 + g_5 V_{ij}^2 + g_6 V_{ij}^3$.

For heat-flux modeling, a modification to the standard gradient diffusion hypothesis (SGDH) is used as the training target. SDGH assumes that the wall-normal heat flux is proportional to the mean temperature gradient:

\begin{equation}
\overline{u_i \theta} = - \alpha_t \partial_i T,
\label{eqn:GDH}
\end{equation}
Usually, $\alpha_t = \mu_t/Pr_t$ represents the turbulent thermal diffusivity, modeled as the ratio of eddy viscosity $\mu_t$ to a constant turbulent Prandtl number $Pr_t = 0.85$. The use of a constant turbulent Prandtl number implies that the turbulence isotropically diffuses heat - an assumption that often is incorrect and this work aims to improve upon. Therefore, here a function is trained to replace the constant $1/Pr_t$ as follows:

\begin{equation}
\begin{aligned}
     \overline{u_i \theta} &= - \alpha^{GEP} \mu_t \partial_i T,\\
     \alpha^{GEP} &= g(A[1:5], I_1, I_2, J_1, J_2, J_3, J_4, J_5 ,+, -, \times)
\label{eqn:trainGDH}
\end{aligned}
\end{equation}
where $J_i$ are temperature invariants. $J_1 = \partial_i T \partial_i T, J_2=\partial_i T S_{ij} \partial_j T, J_3=\partial_i T S_{ij}S_{jk} \partial_k T, J_4=\partial_i T \Omega_{ij}\Omega_{jk} \partial_k T, J_5=\partial_i T \Omega_{ij}S_{jk} \partial_k T$. Therefore, the constants, velocity and temperature invariants, together with a set of mathematical operators, serve as the inputs, while $\alpha^{\mathrm{GEP}}$ is the output for the heat-flux closure development.

\subsubsection{Setup of development and demonstration cases}
% also need to introduce the cost function !!!
Once the mean error and uncertainty were obtained from the surrogate model predictions (see Figure~\ref{fig:surrogateComponents} (c.2)), various combinations of model selection metrics, their corresponding hyperparameters, and selection thresholds were explored during the methodology development phase, as described in Section~\ref{subsection:model selection}.

To identify the recommended setup for the surrogate-augmented CFD-driven framework, four flow cases are considered and grouped according to increasing levels of physical complexity and computation cost. Two development cases that are computationally inexpensive, allows for explore combinations of surrogate model hyperparameters to identify an optimal hyperparameter set. They are a square duct flow, used for turbulence-model training, and a vertical natural convection (VNC) case, used for turbulence and heat-flux model training. To assess the performance and robustness of the framework using the identified hyperparameter configuration, two additional demonstration cases are subsequently considered: a horizontal mixed convection (HMC) configuration and a concentric horizontal annulus (CHA) case. These cases involve higher physical complexity and computational cost and therefore provide a evaluation of the framework’s generality and robustness.

For each flow case, the sources of the high-fidelity datasets, along with the corresponding RANS mesh configurations, numerical setups, and cost-function formulations, are presented in this subsection.

\begin{figure}[t]
\centering
\includegraphics[width=1\linewidth]{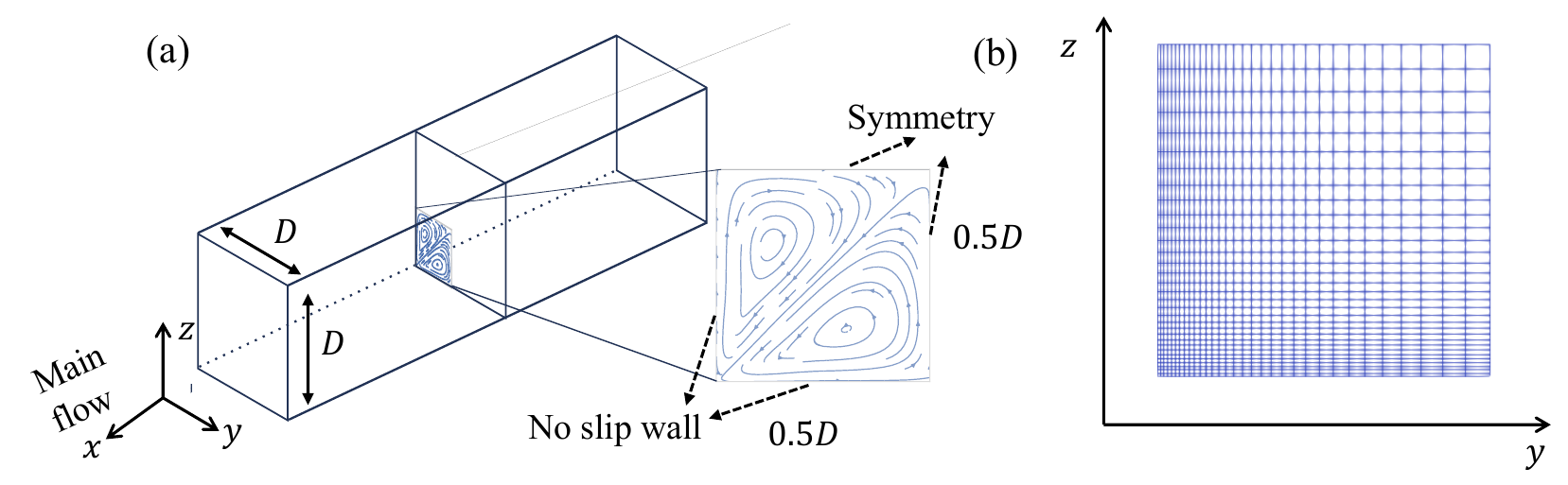}
\caption{Schematic of the duct flow. (a) computational domain with boundary conditions; (b) mesh}\label{fig:duct}
\end{figure}

The duct flow, shown in Figure~\ref{fig:duct}, represents an internal flow with a rectangular cross-section that typically develops mean secondary vortices in the duct corners. However, the RANS approach with the Boussinesq approximation fails to capture these secondary flows, making this case a representative benchmark for turbulence model development. In the present training, the selected case is a square duct with an aspect ratio of 1 and a friction Reynolds number of $Re_{\tau} = 360$. RANS is performed in OpenFOAM v2306 using the incompressible steady-state solver simpleFoam. The computational domain is restricted to one quarter of the rectangular cross-section, with two no-slip walls and two symmetry boundaries applied, as illustrated in Figure~\ref{fig:duct} (a). The corresponding mesh configuration is shown in Figure~\ref{fig:duct} (b). The DNS reference data is from the work of Vinuesa et al.~\cite{vinuesa2014aspect}. In the training setup, the objective is to determine the two coefficients $g_{i} (i=1,2)$ in the additional anisotropic stress term $a_{ij}$ in Equation~\ref{eqn:tauijML}, using two invariants $I_i (i=1,2)$ as inputs. Two cost functions extracted at $y/h=0.5$ are defined: one based on the second velocity component and the other on the third velocity component, with the aim of accurately capturing the secondary flow structures, shown in Table~\ref{table:case}.

\begin{figure}[t]
\centering
\includegraphics[width=1\linewidth]{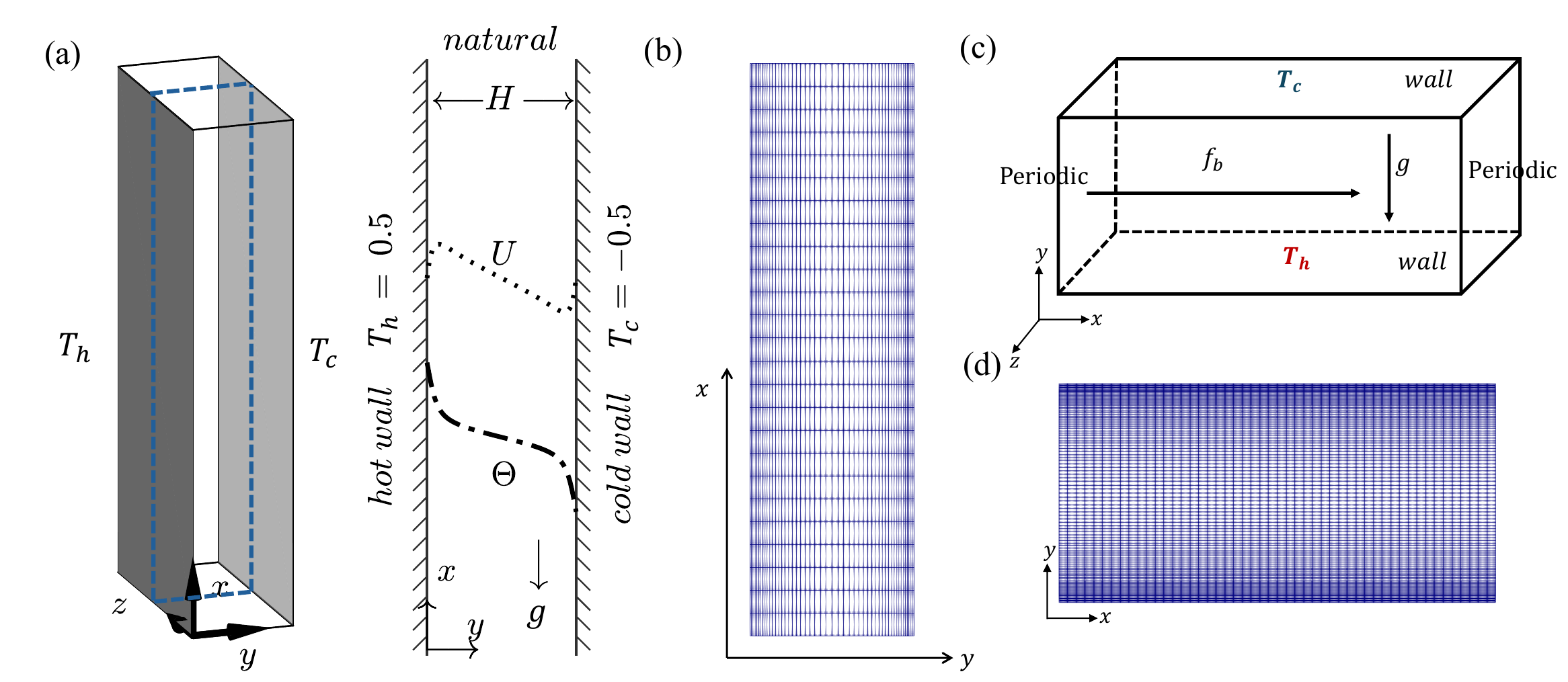}
\caption{Schematic of the vertical natural and horizontal mixed convection flows. (a) and (c) computational domain with boundary conditions; (b) and (d) mesh}\label{fig:vncHmc}
\end{figure}

A series of thermal convection cases is selected to evaluate the framework’s performance in training coupled turbulence and heat-flux models. Statistically, one-dimensional vertical natural convection (VNC) and horizontal mixed convection (HMC) flows are employed as examples. Both VNC and HMC configurations involve fully developed planar channel flows confined between two parallel walls maintained at different temperatures. The flow is driven by buoyancy arising from the imposed temperature gradient, which introduces a strong coupling between the thermal and momentum fields. In the HMC case, an additional constant mean pressure gradient is applied, resulting in combined buoyancy- and shear-driven flow. Figure~\ref{fig:vncHmc} illustrates the computational domains, meshes, and boundary conditions for the VNC and HMC simulations. The development VNC case corresponds to $Ra = 5.4 \times 10^5$ and $Pr = 0.709$, while the HMC case is characterized by $Ra = 10^8, Re_b = 10{,}000$, and $Pr = 1$. The RANS calculations are performed in OpenFOAM v2306 using a self-developed solver for the VNC case (see~\ref{appendix:solver}) and the buoyantBoussinesqSimpleFoam solver for the HMC case.  The high-fidelity datasets are obtained from direct numerical simulations (DNS), with the VNC data from Ng et al.\cite{ng2015vertical} and the HMC data from Pirozzoli et al.\cite{pirozzoli2017mixed}. In the model training framework, the training objectives for the statistically one-dimensional VNC and HMC cases are the coefficients $g_1$ in $a_{ij}$ and $g_4$ in $b_{ij}$ with one invariant $I_1$ used in the turbulence model, and $\alpha^{GEP}$ with two invariants $I_1$ and $J_1$ used in the heat-flux model. The two cost functions are defined based on the mean velocity $J(u)$ and mean temperature $J(T)$ profiles in Table~\ref{table:case}.

\begin{figure}[t]
\centering
\includegraphics[width=0.8\linewidth]{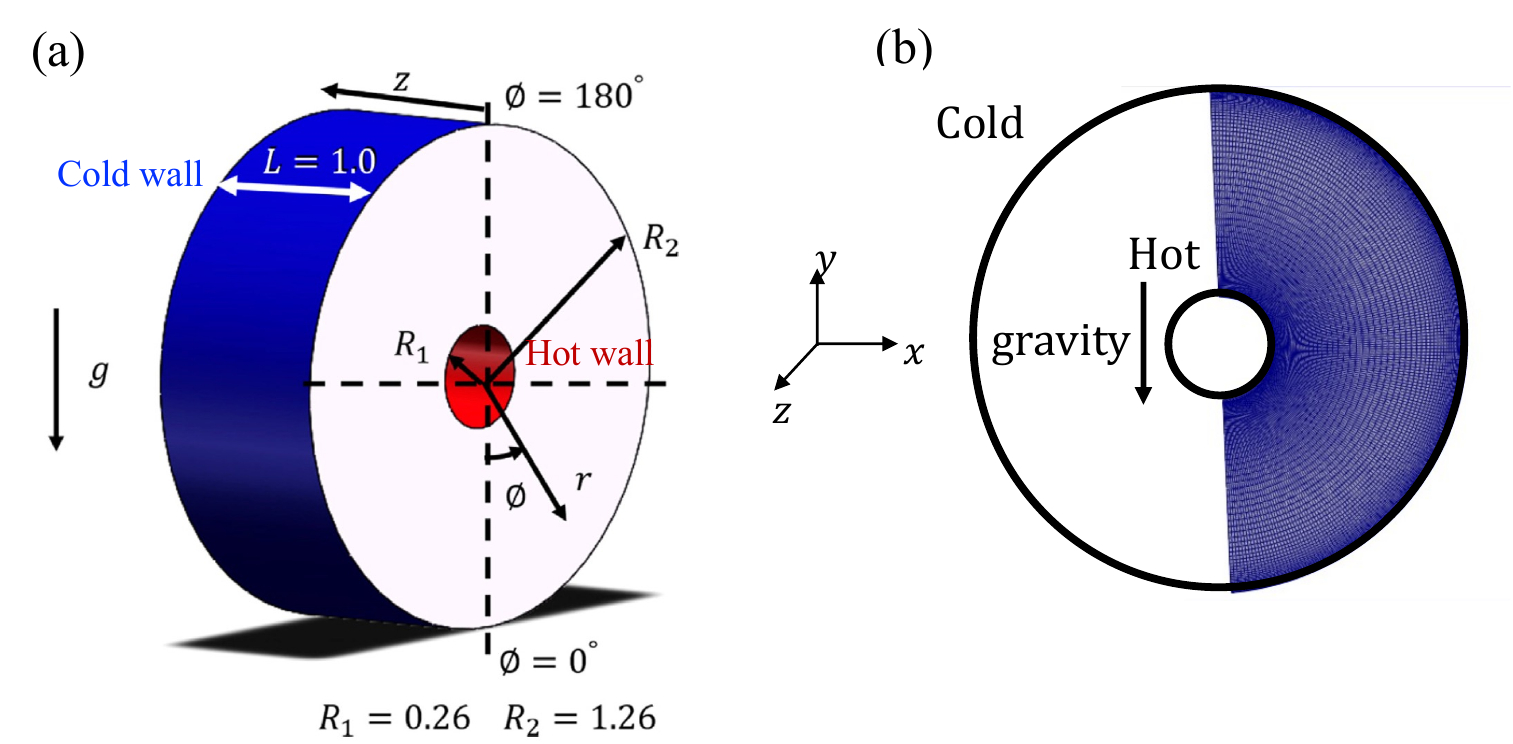}
\caption{Schematic of the horizontal concentric annulus of helium. (a) computational domain with boundary conditions; (b) mesh}\label{fig:cha}
\end{figure}

To advance toward a more realistic and complex configuration involving multiple coupled physical fields, natural convection of helium in a horizontal concentric annulus is selected. Unlike statistically one-dimensional flows, this configuration exhibits distinct flow structures at different angular positions, making it a challenging benchmark for both multi-physics coupled modeling and surrogate modeling techniques. The case chosen corresponds to the highest Rayleigh number, $Ra = 2.38 \times 10^{10}$, available in the DNS reference database of Li et al.~\cite{liyuancha}, with a Prandtl number of Pr = 0.688. The three-dimensional geometric configuration is shown in Figure~\ref{fig:cha} (a), where the inner and outer cylinder radii are $r_{\text{in}} = 0.26$ and $r_{\text{out}} = 1.26$, respectively, yielding an axial wall-to-wall distance of $L = 1$. The mesh configuration is illustrated in Figure~\ref{fig:cha} (b), where only half of the concentric annulus is modeled. The curved surfaces are treated as no-slip walls maintained at different temperatures, while the centerline is assigned a symmetry boundary condition. The RANS calculations are conducted in OpenFOAM v2306 using the buoyantBoussinesqSimpleFoam solver. In the RANS closures training setup, the training goal is 4 expressions with 3 for $g_i (i=1,2,3)$ in $a_{ij}$ in the turbulence model and 1 for $\alpha^{GEP}$ in the heat-flux model. The inputs for the former is two invariants $I_i (i=1,2)$ and three extra input features $N_i (i=1,2,3)$ inspired by the work of Ling et al. \cite{ling2015evaluation}. Specifically, they are selected to  

\begin{table}[t]
\centering
\small
\begin{tabular}{c c c c c c c}
    \hline
  Case & Purposes & Objectives & Inputs & Cost Function & $Pr$ & $Ra$\\
    \hline
  Duct & training & $a_{ij}$ & $I_i$ & $J(\overline{u}_1), J(\overline{u}_2^2+\overline{u}_3^2)$ & - & - \\
  VNC & training & $a_{ij}, R_{ij}, \overline{u'T'}$ & $I_i, J_i$ &$J(u), J(T)$ & 0.709 & 5.4e5\\
  HMC & testing & $a_{ij}, R_{ij}, \overline{u'T'}$ & $I_i, J_i$ & $J(u), J(T)$ & 1 & 10e8\\
  CHA & testing & $a_{ij}, \overline{u'T'}$& $I_i, J_i, N_i$ &$J(u_{180}), J(u_{120})$ &0.688 & 2.38e10\\ 
    \hline
\end{tabular}
\caption{Training and testing flow cases for surrogate-augmented CFD-driven training framework}\label{table:case}
\end{table}

\section{Results and Discussion}
% structure:first show the a large amount of tests on duct and vnc to determine the recommend hyperparameter setup and summarized in a table. Second show the results of two test cases: hmc and cha
This section presents the results for four studied cases—square duct flow, VNC, HMC, and natural convection in a CHA. The discussion focuses on both the performance of the surrogate-augmented CFD-driven training framework compared with the regular CFD-driven approach, and the predictive performance of the trained models for each case. 

\subsection{Development cases: Square duct and Vertical natural convection}
% prove the robust of the algorithm and obtain recommend setup
\subsubsection{Evaluation of Surrogate Model Performance}
As described in the Methodology section (\ref{subsection:model selection}), multiple hyperparameters are involved in configuring the surrogate model. Using the duct and VNC training cases, we aim to identify the recommended configuration for the surrogate model setup. 

The investigation begins with square-duct flow, in which various combinations of hyperparameters are tested. Table~\ref{table:ductHyper} summarizes the investigated hyperparameters of the selection metrics and thresholds employed in the surrogate model to determine which candidate models are selected for CFD re-evaluation. After this analysis, the remaining two hyperparameters — the convergence ratio $\delta$ and the initial selection size $n_i$—are subsequently investigated. In this study, the input mapping from the discrete GEP-generated models to the continuous values used in the surrogate model is achieved by taking the average values of $I_1$ and $I_2$ along a line at half the channel height in the y-direction shown in Figure~\ref{fig:duct}.

\begin{table}[t]
\centering
\footnotesize
\begin{tabular}{c p{2cm} p{2cm} c c p{2cm} c}
    \hline
  Metric & $\beta / \xi$   & $m_{s,r}$ & $m_{s,f}$ & $m_{s,p}$ & $n_i$ & $m_{i,r}$\\
    \hline
  LCB & 1.0,2.0,3.0, 5.0,6.0,8.0 & 0.1, 0.25, 0.5, 0.75, 0.9 & -, 1, 2, 5 & -,1,2 & 5,8,10,12,15 & -, 0.5, 0.75 \\
  EI & 0.01,0.05,0.1, 0.5,1.0,5.0 & 0.05,0.1,0.25, 0.5,0.75,0.9 & -, 1, 2, 5 & -,1,2 & 5,8,10,12,15 & -, 0.5, 0.75  \\
    \hline
\end{tabular}
\caption{Overview of selection metric, threshold  and initial selection hyperparameters investigated via passive training method for the square duct flow}\label{table:ductHyper}
\end{table}

Given the large number of possible hyperparameter combinations, conducting additional CFD-driven training for each configuration would be computationally prohibitive. Therefore, to systematically evaluate the surrogate framework performance without performing further CFD calculations, a database is constructed from 20 regular CFD-driven training runs with different random seeds. In each run, five different random numbers are predefined (as shown in $A[1:5]$ in Equation~\ref{eqn:tauijML}), and the resulting datasets are subsequently utilized for passive training. In the passive training process, the influence of employing a surrogate model and different selection strategies—namely, the initial selection, selection metric, and threshold settings—is emulated using the existing CFD-driven training data rather than running CFD calculations. Each generation of the emulated training loop consists of three steps. (1) selecting candidate models from the initial model population, (2) adding the corresponding CFD data of the selected models to the training set, and (3) retraining the surrogate model.

\begin{figure}[htp]
    \centering
    \includegraphics[width=0.5\linewidth]{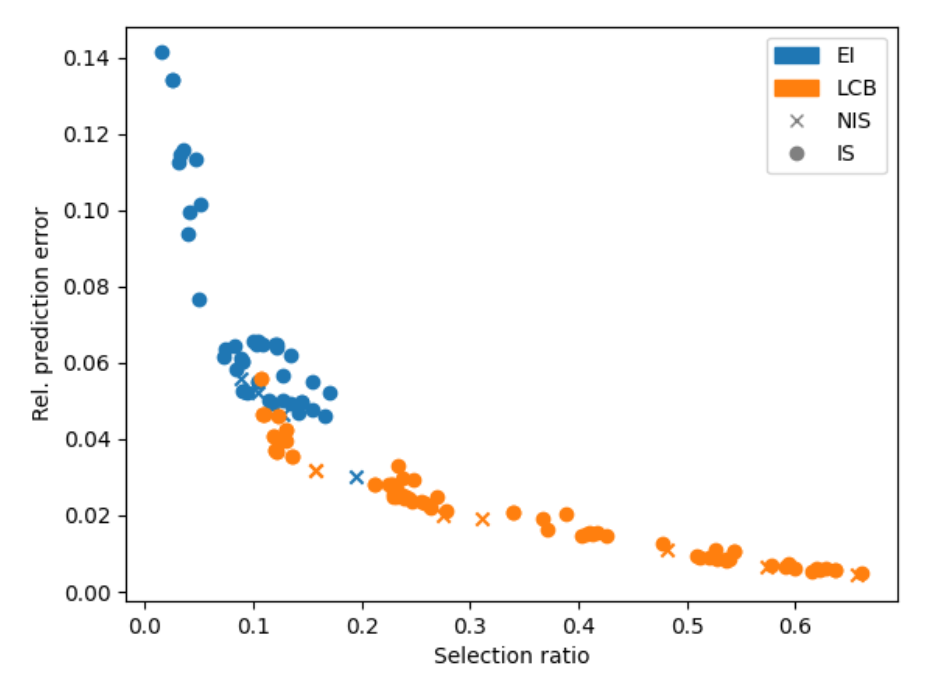}
    \caption{Passive training performance of varying selection metric, threshold and initial selection parameters on the duct flow}
    \label{fig:ductSorrgateHyper}
\end{figure}

To identify the optimal surrogate model configuration and selection process, a total of 59 different combinations of selection metrics and threshold settings, based on the hyperparameters listed in Table~\ref{table:ductHyper}, were tested. The surrogate models' relative prediction error with respect to the true CFD evaluations, along with the selection ratio (i.e., the proportion of candidate models selected for CFD re-evaluation relative to the total evaluations in the regular CFD-driven training), are plotted in Figure~\ref{fig:ductSorrgateHyper}.

Based on the passive training results, the parameter settings achieving the optimal trade-off value in Figure~\ref{fig:ductSorrgateHyper} were selected for further validation using surrogate-augmented CFD-driven training:

\begin{enumerate}
    \item V1: EI, $\xi=0.1, m_{s,r}=0.9, n_i=15$
    \item V2: EI, $\xi=0.1, m_{s,r}=0.05, m_{s,f}=1, n_i=10, m_{i,r}=0.5$
    \item V3: LCB, $\beta=5.0, m_{s,r}=0.25, m_{s,f}=1, n_i=8, m_{i,r}=0.5$
    \item V4: EI, $\xi=1.0, m_{s,r}=0.1$
    \item V5: LCB, $\beta=5.0, m_{s,r}=0.25, m_{s,p}=1, n_i=8, m_{i,r}=0.5$
    \item V6: LCB, $\beta=5.0, m_{s,r}=0.25, m_{s,f}=2$
    \item V7: LCB, $\beta=3.0, m_{s,r}=0.25, m_{s,p}=1, n_i=10, m_{i,r}=0.5$
    \item V8: LCB, $\beta=3.0, m_{s,r}=0.25, m_{s,p}=1$
    \item V9: LCB, $\beta=5.0, m_{s,r}=0.25, m_{s,p}=2$
\end{enumerate}

To compare the surrogate model performance for the multi-objective optimization of the duct flow problem $(J(\overline{u}_1), J(\overline{u}_2^2 + \overline{u}_3^2))$ across the various setups V1–V9, two indicators are considered: the number of CFD calculations and a metric called hypervolume coverage, which quantifies how much of the optimal region is occupied by the obtained Pareto front.
Hypervolume coverage measures the multi-dimensional volume between the Pareto front and a reference point. Here, the maximum values of each objective on the Pareto front are used as the reference point. A larger hypervolume coverage, therefore, indicates better overall performance. 

\begin{figure}[htp]
    \centering
    \includegraphics[width=1.0\linewidth]{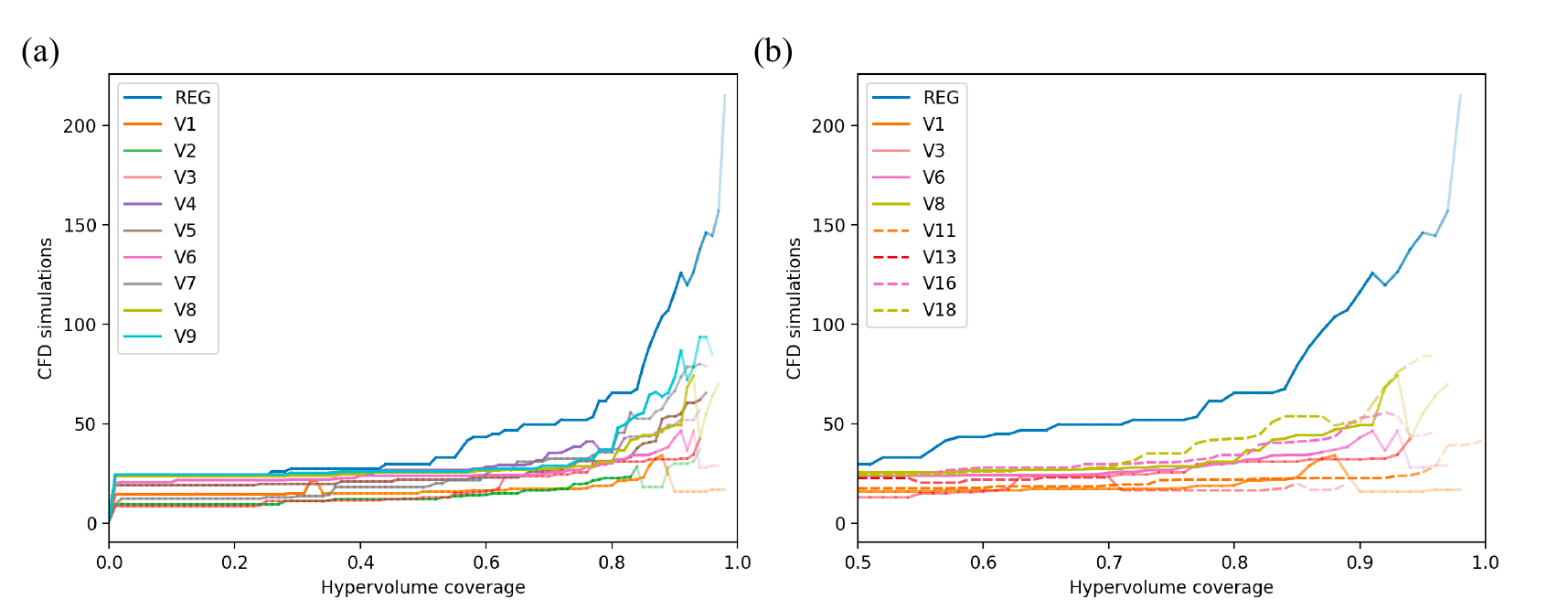}
    \caption{Computational costs, i.e. number of CFD evaluations, to achieve hypervolume coverage without (a) and with (b) convergence weight, averaged over 4 validation initializations for multi-objective optimization problem of the square duct flow. Opacity of lines represents ratio of training runs reaching the respective fitness level.}
    \label{fig:ductv19}
\end{figure}

The results for V1–V9 are presented in Figure~\ref{fig:ductv19} (a). Compared with the baseline CFD-driven training (highlighted in blue), all surrogate-augmented CFD-driven training cases achieve comparable hypervolume coverage with substantially fewer CFD evaluations. A closer examination shows that V1, V3, V6, and V8 achieve larger hypervolume coverage values, approaching unity, than the other configurations. To further improve robustness, a convergence weight of $\delta = 0.5$ was introduced in Equation~\ref{eqn:CW} for these four configurations to help the surrogate model exclude non-converged samples, resulting in additional configurations V11, V13, V16, and V18, shown in Figure~\ref{fig:ductv19} (b). Among these, V11, V8, and V6 exhibit higher hypervolume coverage, whereas V13 yields a significantly lower CFD cost. Thus, the V6, V8, V11, and V13 setups are recommended hyperparameter combinations. Further selection and validation will be carried out on the VNC case, where both turbulence and heat-flux models are trained simultaneously.

Building on the four recommended surrogate hyperparameter setups (V6, V8, V11, and V13) identified from the square duct flow case, we further test these configurations on a more complex optimization problem, where both the turbulence and the heat-flux models are simultaneously improved for the statistically one-dimensional VNC case, to examine whether the recommended hyperparameter set can be further refined. For convenience, the four best-performing hyperparameter configurations from the duct-flow case, V6, V8, V11, and V13, are relabeled as V1–V4, respectively. In addition, four new configurations (V5–V8) are introduced to examine different values of the convergence ratio $\delta$. The specific setups are as follows: 

\begin{enumerate}
    \item V1: LCB, $\beta=5, m_{s,r}=0.25, m_{s,f}=2$
    \item V2: LCB, $\beta=3.0, m_{s,r}=0.25, m_{s,p}=1$
    \item V3: EI, $\xi=0.1, m_{s,r}=0.9, n_i=75, \delta=0.5$
    \item V4: LCB, $\beta=5.0, m_{s,r}=0.25, m_{s,f}=1, n_i=40, m_{i,r}=0.5,\delta=0.5$
    \item V5: LCB, $\beta=5, m_{s,r}=0.25, m_{s,f}=2, \delta=0.5$
    \item V6: LCB, $\beta=3.0, m_{s,r}=0.25, m_{s,p}=1, \delta=0.5$
    \item V7: EI, $\xi=0.1, m_{s,r}=0.9, n_i=75, \delta=0.1$
    \item V8: LCB, LCB, $\beta=5.0, m_{s,r}=0.25, m_{s,f}=1, n_i=40, m_{i,r}=0.5,\delta=0.75$
\end{enumerate}

The input mapping for the surrogate model is obtained using the averaged values of $I_1$ and $J_1$ across the distance between the two differentially heated walls shown in Figure~\ref{fig:vncHmc} (a), based on the baseline RANS results. For the statistically one-dimensional VNC flow, the turbulence-model training produces one coefficient, $g_1$, associated with the first tensor basis $\mathbf{V}_{ij}^1$ in Equation~\ref{eqn:tauijML}. Based on previous VNC training experience, the coefficient $g_4$ of the first tensor in the artificial turbulence-production term R (Equation~\ref{eqn:extra}) is also included, together with the heat-flux term $\alpha^{GEP}$ in Equation~\ref{eqn:trainGDH}, resulting in three models being trained simultaneously in the multi-objective optimization.

\begin{figure}[htp]
    \centering
    \includegraphics[width=1.0\linewidth]{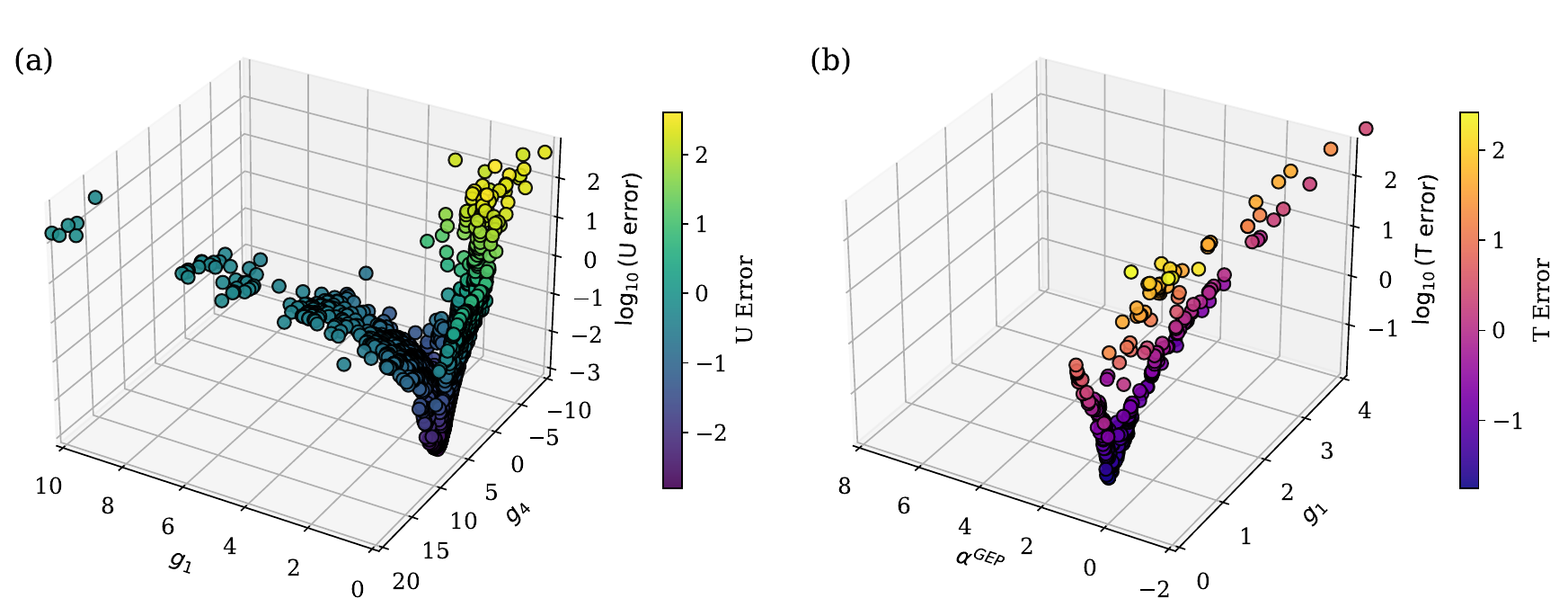}
    \caption{Mapping the regular CFD-driven trained models to predicted errors: (a) velocity error using the GEP-generated turbulence model with an additional turbulence production term; (b) temperature error using the GEP-generated turbulence and heat flux models.}
    \label{fig:surMap}
\end{figure}

After performing the regular CFD-driven training for the VNC turbulence and heat-flux model development, a preliminary test is conducted prior to the surrogate-augmented CFD-driven training to evaluate whether the surrogate model can effectively map the GEP-trained models to the CFD-computed errors. Figure \ref{fig:surMap} (a) shows all GEP-generated turbulence models, including $g_1$ and $g_4$, plotted against the corresponding velocity errors on a logarithmic scale to better illustrate the distribution. Figure \ref{fig:surMap} (b) presents the GEP-generated turbulence model $g_1$ and the heat flux model $\alpha^{GEP}$ plotted against the corresponding temperature errors, also scaled logarithmically. Both plots exhibit distinct regions of local minima, suggesting that the surrogate model can potentially capture the relationship between the trained models and the observed errors.

\begin{figure}[htp]
    \centering
    \includegraphics[width=1.0\linewidth]{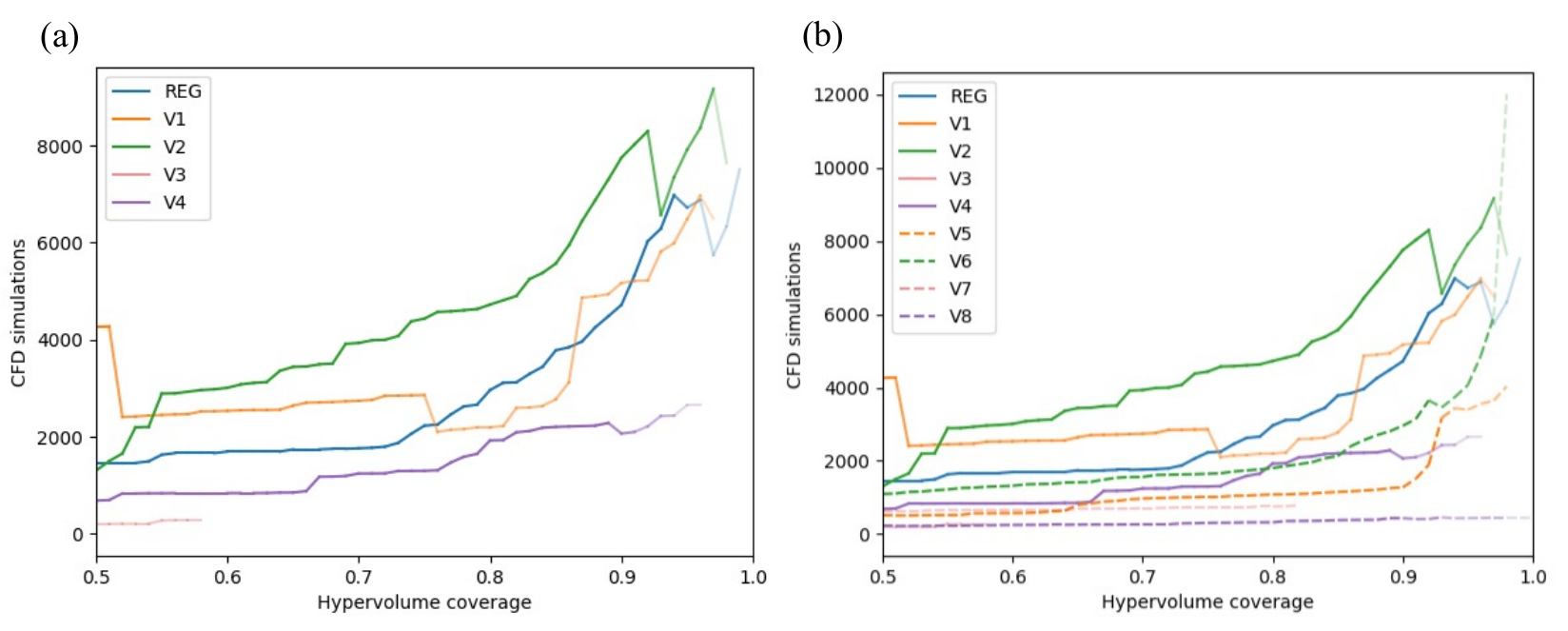}
    \caption{Computational costs, i.e. number of CFD evaluations, to achieve hypervolume coverage (a) V1-V4 (b) tuning convergence weight averaged over 4 validation initializations for multi-objective optimization problem of the VNC. Opacity of lines represents ratio of training runs reaching the respective fitness level.}
    \label{fig:vncv18}
\end{figure}

After gaining confidence in the surrogate model’s performance on the coupled model predictions shown in Figure~\ref{fig:surMap}, we proceeded to apply the hyperparameter setups V1–V8 to the VNC case, as shown in Figure~\ref{fig:vncv18}. Figure~\ref{fig:vncv18} (a) corresponds to the best-performing setups identified from the duct-flow tests, while Figure~\ref{fig:vncv18} (b) includes additional configurations obtained by introducing a convergence weight of $\delta = 0.5$ for V1 and V2 (yielding V5 and V6), and by decreasing and increasing the convergence weight for V3 and V4, resulting in V7 and V8, respectively. In Figure~\ref{fig:vncv18} (a), V4 emerges as the best setup, as it requires fewer CFD simulations compared with V1 and V2, while V3 only reaches a limited hypervolume coverage, with relatively few CFD simulations concentrated in regions of high EI. After assigning or adjusting the convergence ratio $\delta$, as shown in Figure~\ref{fig:vncv18} (b), all configurations V5–V8 show improved performance compared with V1–V4, confirming the importance of the convergence ratio $\delta$ hyperparameter. Among all cases, V8 demonstrates the best overall performance, achieving a high hypervolume coverage with a relatively small number of required CFD simulations. In the remaining two test cases, V8 is therefore adopted as the default hyperparameter setup for the surrogate model within the surrogate-augmented CFD-driven training framework.

\begin{figure}[htp]
    \centering
    \includegraphics[width=0.5\linewidth]{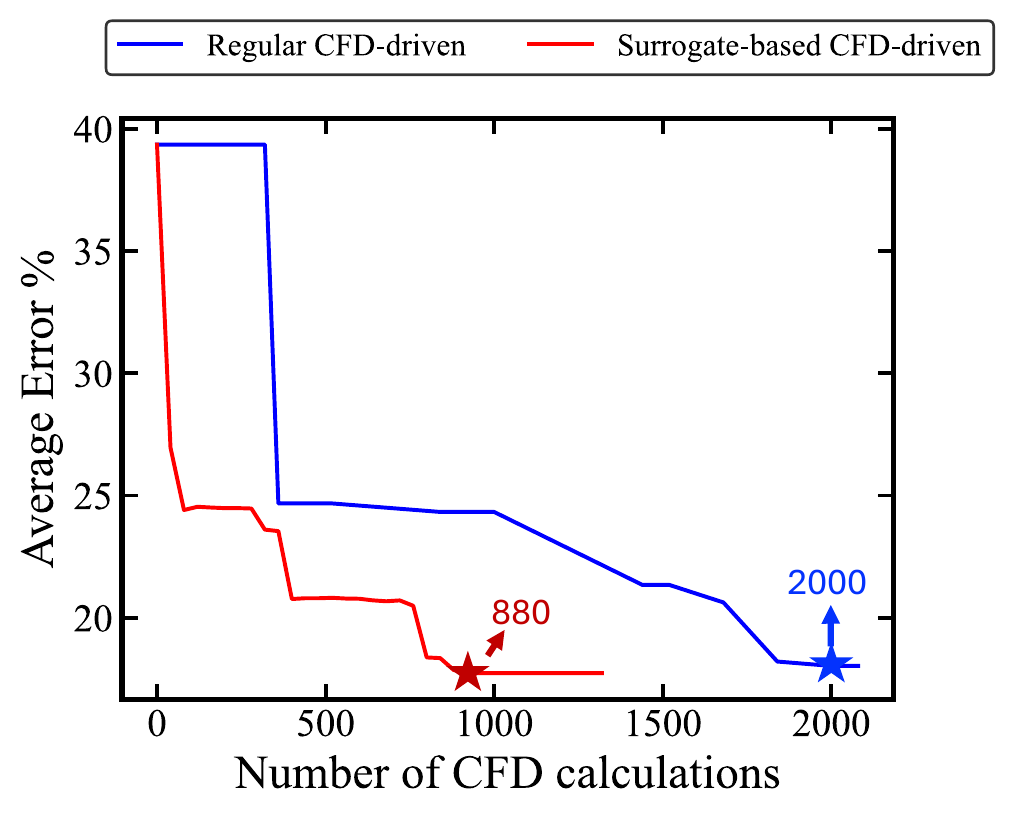}
    \caption{Error evolution in regular and surrogate-based CFD-driven training for the VNC case}
    \label{fig:eff_vnc}
\end{figure}

The most straightforward way to evaluate training efficiency is to compare the average prediction error with the number of CFD evaluations required to reach that error level. As shown in Figure~\ref{fig:eff_vnc}, the surrogate-based framework achieves the same minimum average error with only 880 CFD evaluations, compared to 2000 required by the regular approach. This results in a training cost reduction of $(2000-880)/2000=56\%$. 

\subsubsection{Performance of CFD-Driven Trained Models}
The above results focus solely on analyzing the surrogate model hyperparameter settings and the associated reduction in training cost relative to the standard CFD-driven training framework. This subsection presents the CFD-driven trained models with the smallest errors for the square duct and VNC flows, along with their explicit formulations in Appendix~\ref{appendix:models}.

\begin{figure}[htp]
    \centering
    \includegraphics[width=0.8\linewidth]{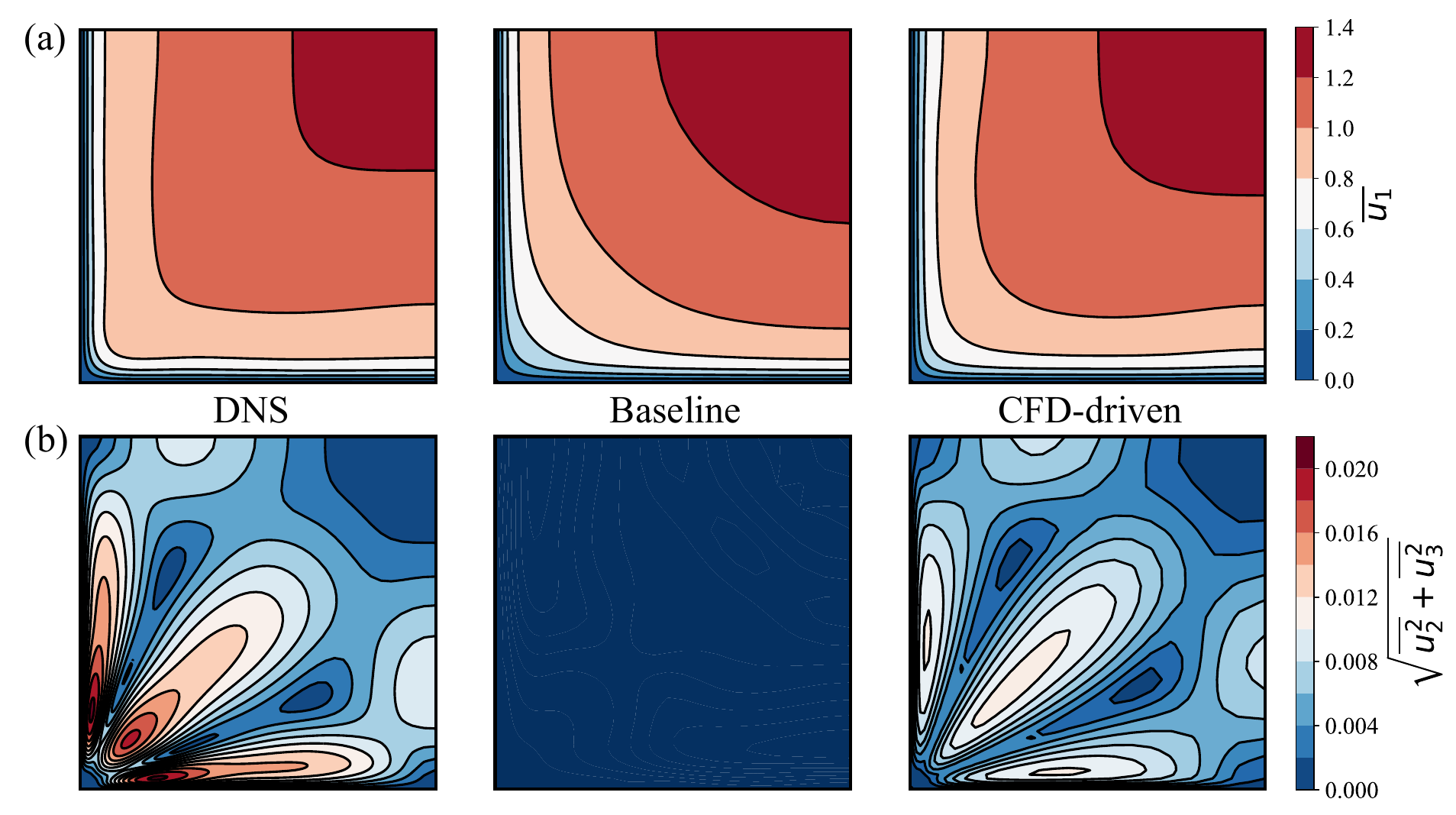}
    \caption{Comparison velocity contours between DNS, baseline RANS and CFD-driven trained model: (a) $\overline{u_1}$; (b) $\sqrt{\overline{u_2^2}+\overline{u_3^2}}$.}
    \label{fig:ducttrained}
\end{figure}

The performance of the CFD-driven trained turbulence model that achieved the minimal error on the square duct is presented in Figure~\ref{fig:ducttrained}, demonstrating the effectiveness of the CFD-driven training approach for turbulence model development. Figure~\ref{fig:ducttrained} (a) compares the streamwise velocity component, while Figure~\ref{fig:ducttrained} (b) compares the magnitude of the spanwise and wall-normal velocity components obtained from DNS, the baseline RANS, and the CFD-driven trained model. The contours from the CFD-driven trained model capture the secondary flow structures and show better agreement with the DNS results than those from the baseline RANS.

\begin{figure}[htp]
    \centering
    \includegraphics[width=0.8\linewidth]{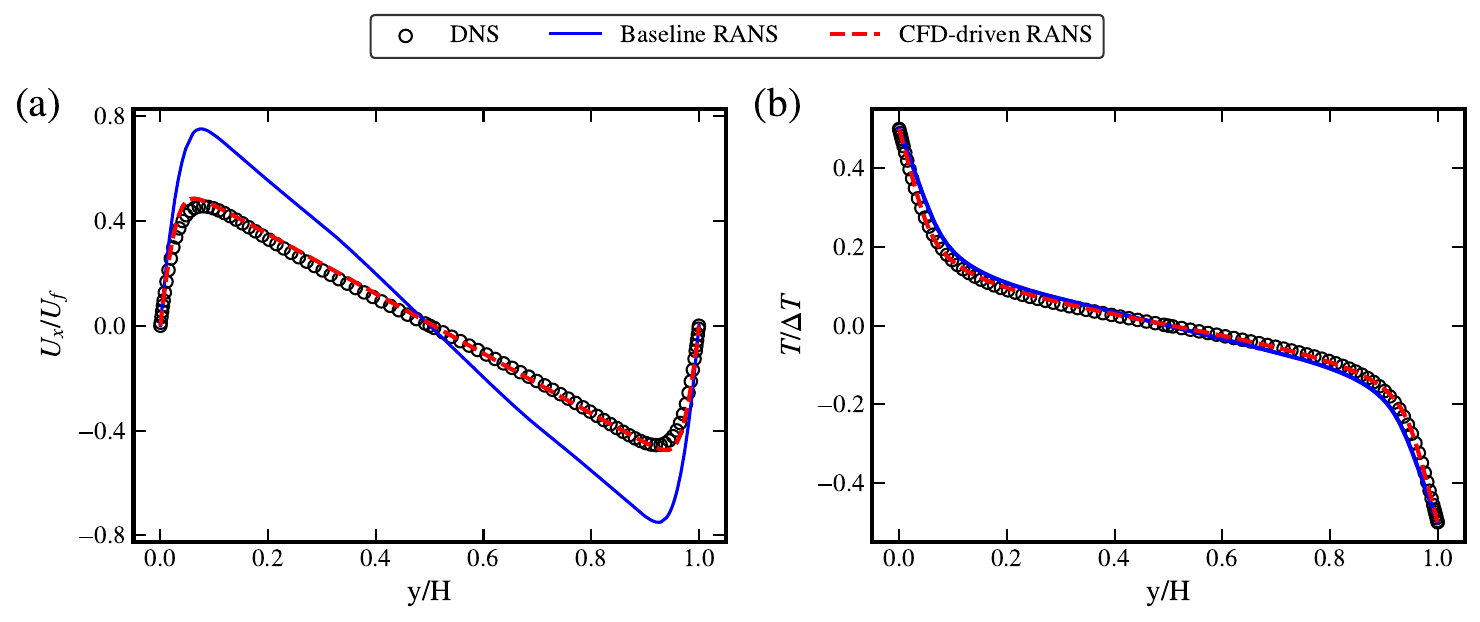}
    \caption{Performance of the CFD-driven trained models on cost function components: (a) mean velocity and (b) mean temperature profiles across the channel.}
    \label{fig:vncmean}
\end{figure}

The performance of the CFD-driven trained models on the VNC case for the two training objectives, velocity and temperature profiles across the channel height, is presented in Figure~\ref{fig:vncmean}. Compared to the baseline models (blue solid line), the set of CFD-driven trained models (red dashed line) defined by the explicit expressions in \ref{appendix:models} exhibits good agreement with the DNS results (black circles).

\begin{figure}[htp]
    \centering
    \includegraphics[width=1.0\linewidth]{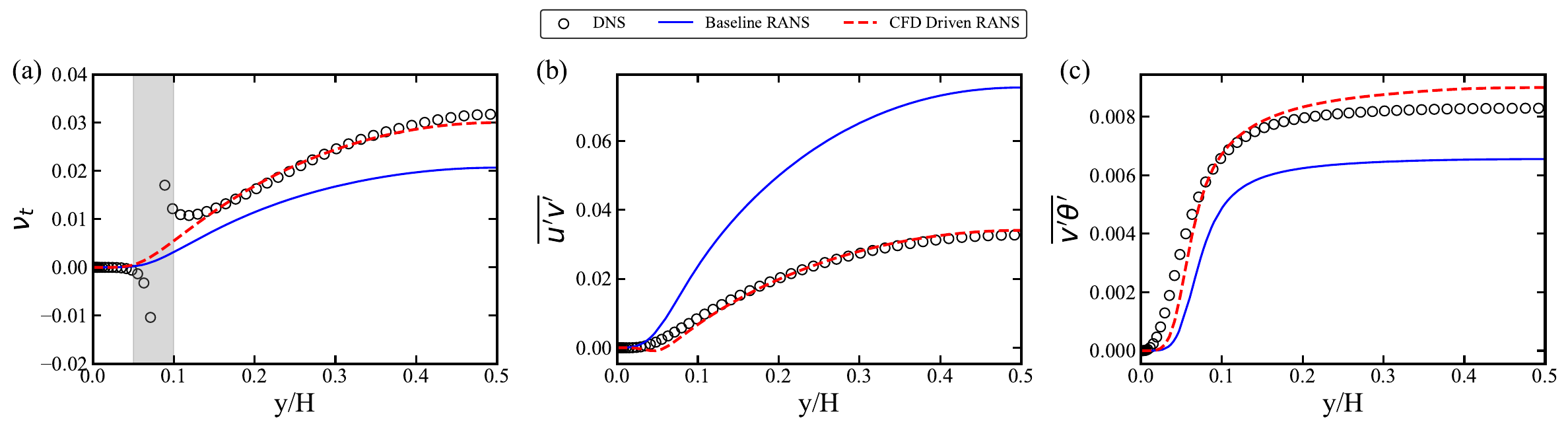}
    \caption{Performance of the CFD-driven trained model on non cost function components: (a) turbulence viscosity, (b) the Reynolds shear stress and (c) the wall-normal heat flux.}
    \label{fig:vncSecond}
\end{figure}

For this statistically one-dimensional mean flow, the governing equations involve only two components of the Reynolds stress tensor and the heat flux vector. With the availability of DNS data, the accuracy of prediction for secondary quantities, such as the turbulence viscosity, Reynolds shear stress, and wall-normal heat flux, can be evaluated, as shown in Figure~\ref{fig:vncSecond}. Based on the Boussinesq assumption, the turbulence viscosity from DNS is obtained using $-\overline{u^{'}v^{'}}=\nu_t\frac{dU}{dy}$ and compared with the turbulence viscosity predicted by the baseline and CFD-driven trained RANS models (Figure~\ref{fig:vncSecond} a).
For the heat flux model evaluation, the wall-normal heat flux is computed using SGDH, $-\overline{v^{'}\theta^{'}}=\alpha_t\frac{dT}{dy}=\frac{\nu_t}{Pr_t}\frac{dT}{dy}$ and compared directly with DNS results (Figure~\ref{fig:vncSecond} c). The results from the CFD-driven RANS model show much closer agreement with DNS than those of the baseline model, even for secondary quantities not included in the training cost function. This consistent improvement arises because the model training retains rich physical information within the CFD solver, rather than merely fitting to a limited set of target quantities.

The Nusselt number, defined as $Nu=(H/\Delta T)\frac{d T}{dy} |_w$, characterizes the ratio of convective to conductive heat transfer and is a key parameter of significant industrial relevance. It is therefore evaluated and compared using both high- and low-fidelity datasets. For this VNC case, the DNS predicts a Nusselt number of 5.37, whereas the baseline RANS model yields 4.33, corresponding to a $19.4\%$ relative error. In contrast, the CFD-driven trained model predicts 4.86, achieving more than a $50\%$ reduction in error ($9.5\%$), thereby demonstrating improved capability to capture near-wall thermal behavior.

\subsection{Demonstration cases: horizontal mixed convection and natural convection of concentric horizontal annulus}
\subsubsection{Evaluation of Surrogate Model Performance}
With the optimal surrogate model configuration determined in the previous section, the surrogate-augmented CFD-driven training framework is now tested on more complex flow cases. The first case considers HMC, where the flow is driven not solely by buoyancy but by the combined effects of buoyancy and an imposed pressure gradient. DNS data for velocity and temperature profiles across a horizontal channel at $Ra = 10^8$ and three bulk Richardson numbers ($Ri_b = 10, 1, 0.1$) are taken from Pirozzoli et al. \cite{pirozzoli2017mixed}. The second test case extends the coupled model training to higher-dimensional mean flows, focusing on natural convection in a concentric horizontal annulus at a high Rayleigh number of $Ra = 2.38\times10^{10}$. The corresponding DNS reference data are taken from \cite{liyuancha}. 

\begin{figure}[htp]
    \centering
    \includegraphics[width=1.0\linewidth]{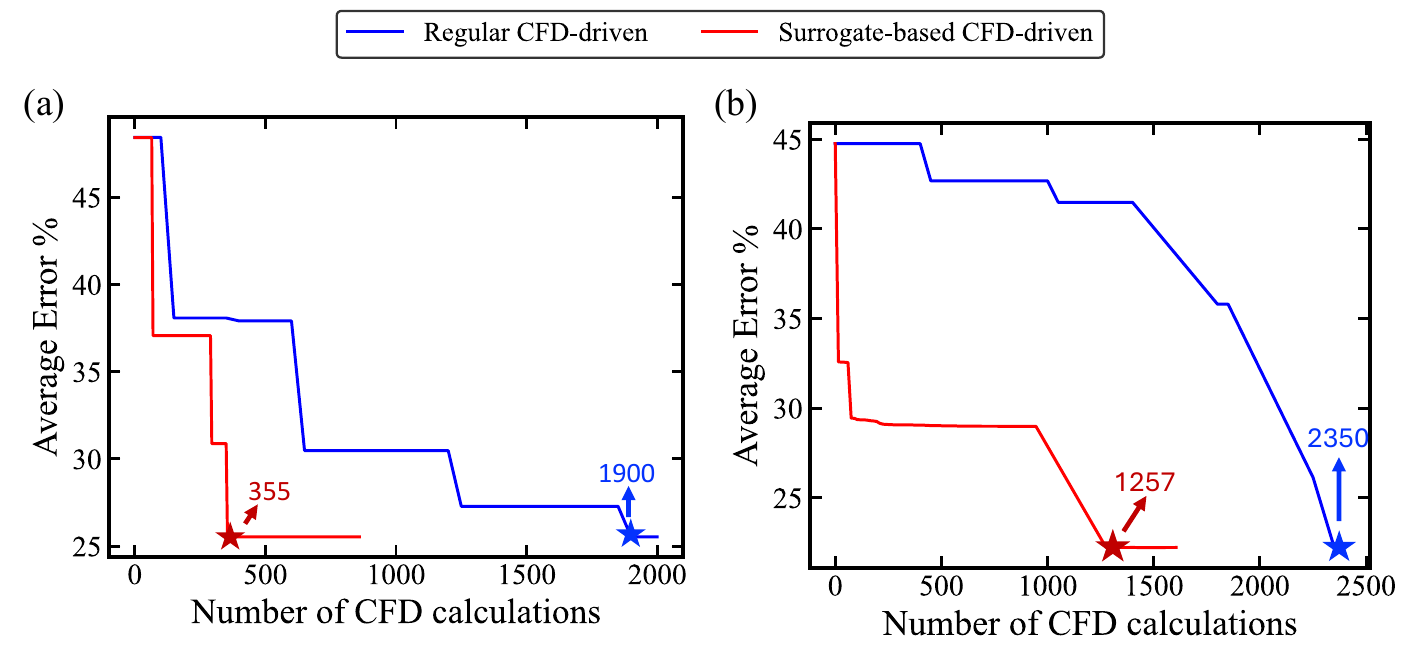}
    \caption{Error evolution in regular and surrogate-augmented CFD-driven training (a) HMC, (b) CHA.}
    \label{fig:hmcChaSur}
\end{figure}

To quantitatively evaluate the training efficiency, Figure~\ref{fig:hmcChaSur} presents the evolution of the CFD error for both the regular and surrogate-augmented training approaches. As shown in Figure~\ref{fig:hmcChaSur} (a), for the HMC case, the surrogate-based model attains the same level of error reduction after 355 CFD evaluations, whereas the regular framework requires 1900 evaluations—indicating an $81.3\%$ improvement in training efficiency. Figure~\ref{fig:hmcChaSur} (b) shows the corresponding results for the CHA case. To achieve a $90\%$ reduction in average error relative to the baseline, the surrogate-based training requires only 1,257 CFD evaluations, compared with 2,350 for the regular framework, corresponding to a $46.5\%$ improvement in efficiency.

In summary, with the recommended hyperparameter configuration for the surrogate model, the proposed framework achieves a substantial improvement in training efficiency across both framework development cases and the two additional demonstration cases, which progressively increase in coupling complexity, flow dimensionality, and Rayleigh number.

\subsubsection{Performance of CFD-Driven Trained Models}
\begin{figure}[htp]
    \centering
  \includegraphics[width=1.0\linewidth]{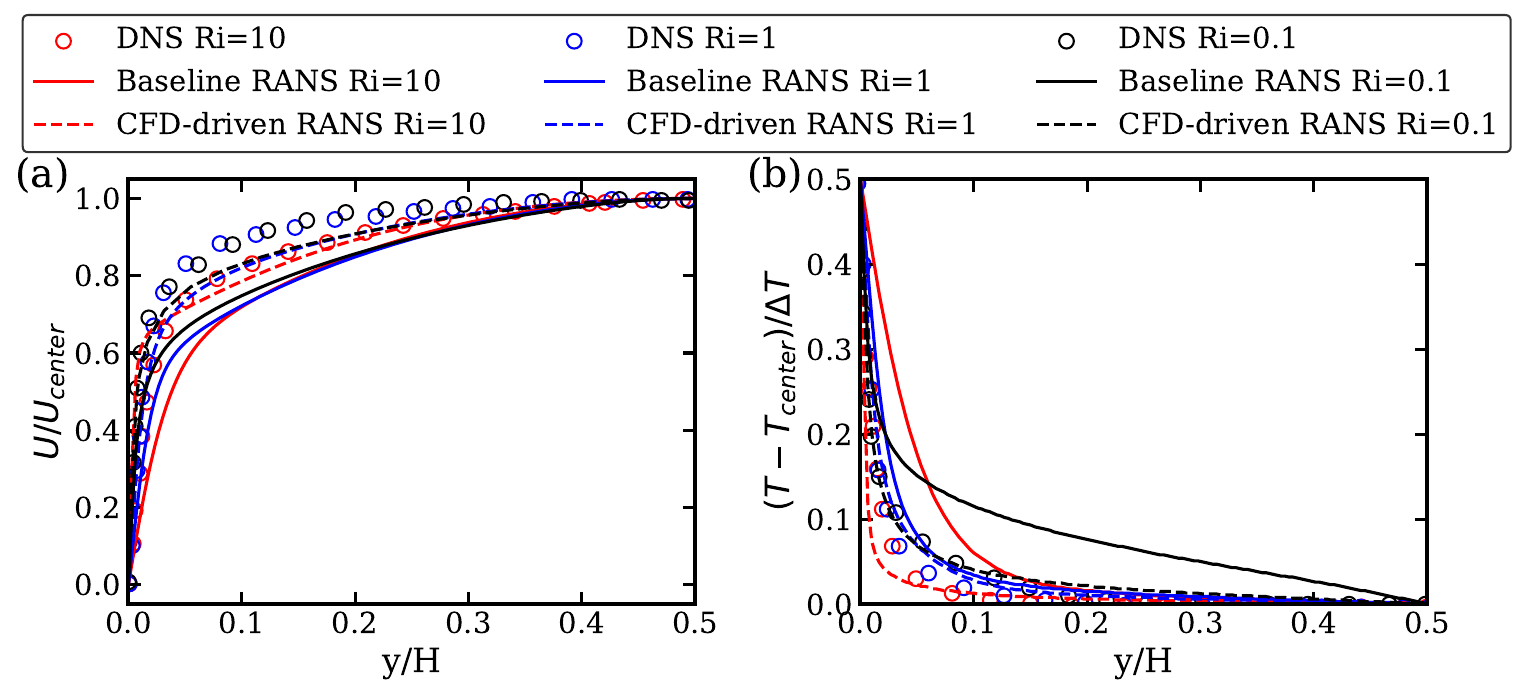}
    \caption{Comparison of mean flow quantities between DNS, baseline RANS and CFD-driven RANS (a) velocity profile, (b) temperature profile}
    \label{fig:hmcut}
\end{figure}

We now examine the performance of the CFD-driven trained model on the two demonstration cases: HMC and CHA. We only train the $Ra=10^8$ and $Ri=0.1$ case. Although DNS turbulent quantities are limited to $Ra=10^8$ in the paper, mean flow quantities, including velocity and temperature, are available for the same Rayleigh number but across varying Richardson numbers. This allows us to assess the generalizability of the GEP-generated model by testing it on cases with the same Ra but different Ri. We begin by examining the performance of the CFD-driven trained HMC model by comparing its mean flow predictions, as shown in Figure~\ref{fig:hmcut}. Figure~\ref{fig:hmcut} shows that the CFD-driven RANS model, trained on a single case, produces mean velocity and temperature predictions that agree better with the DNS data across all Richardson number cases than baseline RANS results.

Moreover, since the Nusselt number is a key quantity of industrial interest and is available from DNS results, it has been calculated using both the baseline and CFD-driven RANS models for all three Richardson-number cases. The corresponding values and errors are summarized in Table~\ref{tab:hmcNu}. The results again show that even when only high-fidelity mean-flow quantities are used during model training, their corresponding gradients can be significantly improved.

\begin{table}[!ht]
    \centering
    \footnotesize
    \caption{Comparison of Nusselt number between DNS, baseline RANS and CFD-driven RANS}
    \label{tab:hmcNu}
    \begin{tabular}{c|c|c|c|c|c}
        \hline
         Cases & DNS & Baseline & Baseline Error & CFD-driven & CFD-driven Error \\
         \hline
        RunRa8Ri10 & 27.672 & 8.372 & $69.746\%$ & 27.807 &  $0.48\%$ \\       
        RunRa8Ri1 & 25.443 & 16.744 &$34.190\%$ &26.312&  $3.303\%$\\
        RunRa8Ri0.1 & 45.584 & 33.787 & $25.880\%$ &43.056& $5.56\%$\\
        \hline
    \end{tabular}
\end{table}

\begin{figure}[htp]
    \centering
  \includegraphics[width=1.0\linewidth]{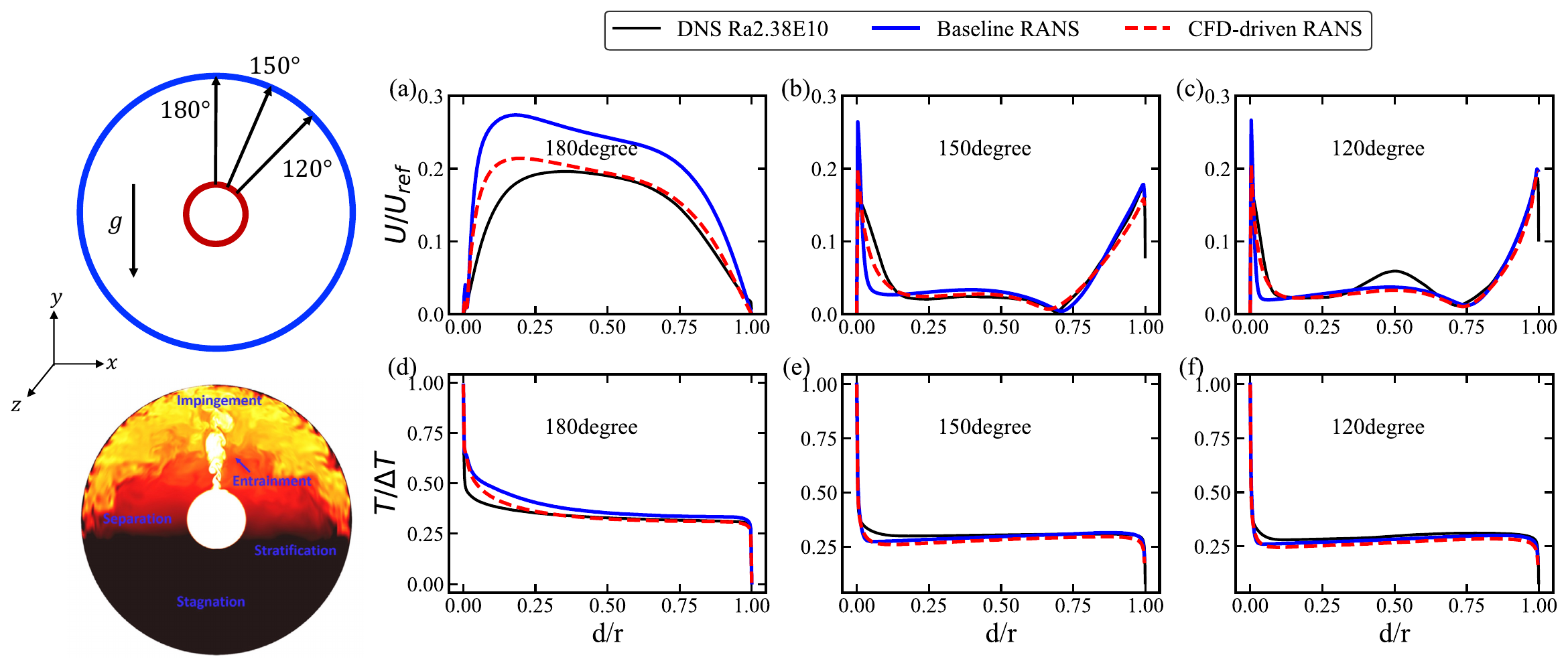}
    \caption{Comparison of mean velocity ((a)-(c)) and temperature profiles ((d)-(e)) between DNS and baseline RANS at three representative angular locations.}
    \label{fig:chaCFDdriven}
\end{figure}

For the CHA case with $Ra = 2.38\times10^{10}$, the two cost function components, velocity and temperature profiles at $180^{\circ} and 120^{\circ}$, as well as the non–cost-function profiles at $150^{\circ}$, are shown in Figures~\ref{fig:chaCFDdriven}. Figures~\ref{fig:chaCFDdriven} (a)–(c) compare the velocity profiles predicted by the DNS, baseline RANS, and CFD-driven trained RANS models, while Figures~\ref{fig:chaCFDdriven} (d)–(f) present the corresponding temperature profiles. In both cases, the coordinate d/r increases from zero (the hot inner wall) to unity (the cold outer wall). At $180^{\circ}$, corresponding to the buoyancy-induced jet region, the baseline RANS model exhibits the most considerable discrepancies in both velocity and temperature compared to DNS, whereas the CFD-driven model shows substantial improvement. A closer examination of the entrainment region (Figures~\ref{fig:chaCFDdriven} (b), (c), (e), and (f)) reveals that the predicted velocity near the cold wall agrees better with DNS than that near the hot wall, with a similar trend observed in the temperature profiles. 

\begin{figure}[htp]
    \centering
  \includegraphics[width=1.0\linewidth]{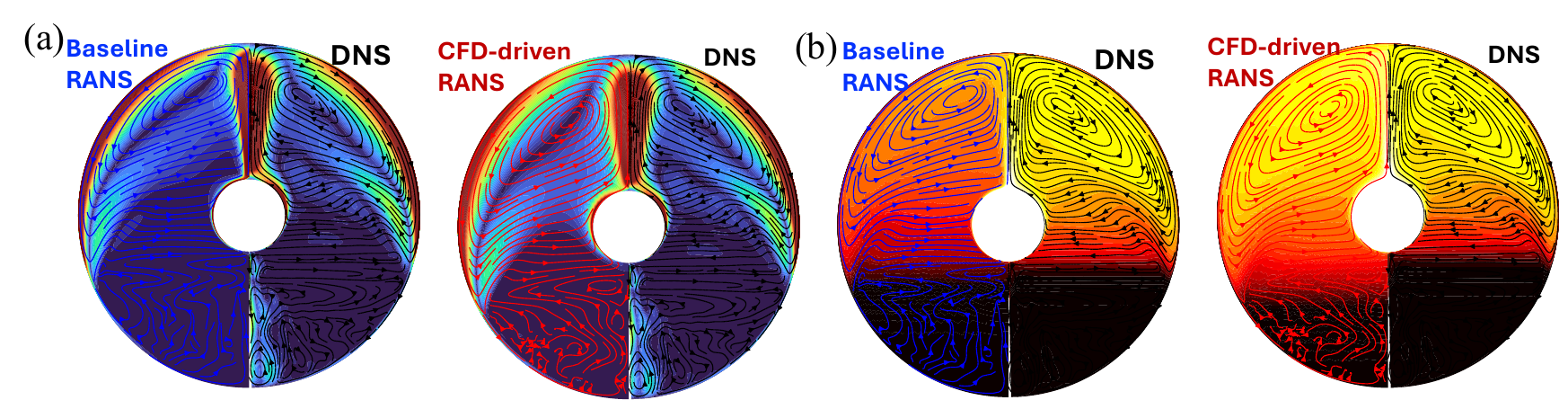}
    \caption{Comparison of mean flow quantities between DNS, baseline RANS, and CFD-driven RANS; (a) velocity magnitude contours with streamline, (b) temperature contours with streamline}
    \label{fig:chaTemVelo}
\end{figure}

To better visualize the mean flow field at various angular positions, velocity-magnitude contours with streamlines are shown in Figure~\ref{fig:chaTemVelo} (a), and temperature contours with streamlines are shown in Figure~\ref{fig:chaTemVelo} (b). The right half of each figure presents the DNS results, while the left half shows the predictions from the baseline RANS and CFD-driven models. In Figure~\ref{fig:chaTemVelo} (a), a buoyancy-driven jet can be clearly observed rising from the inner hot wall and impinging on the cold outer wall. The flow then turns laterally due to outer-wall blockage, forming two large vortices characterized by reduced near-wall velocity and the entrainment of surrounding fluid into the buoyant jet. The CFD-driven model accurately predicts the vortex-center height, consistent with DNS, and significantly improves predictions of the jet’s width and length compared with the baseline model. In Figure~\ref{fig:chaTemVelo} (b), the temperature field predicted by the CFD-driven model exhibits a distinct stratification similar to that observed in DNS, with high temperatures concentrated near the inner hot wall and decreasing gradually toward the outer cold wall.

\begin{figure}[htp]
    \centering
  \includegraphics[width=1.0\linewidth]{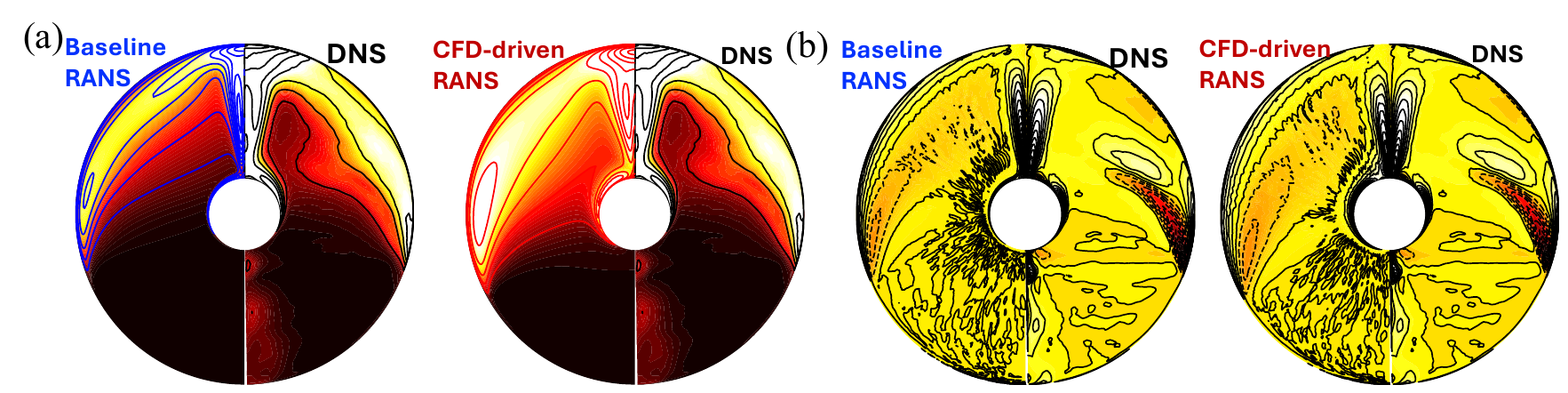}
    \caption{Comparison of turbulence quantities between DNS, baseline RANS, and CFD-driven RANS:
(a) TKE; (b) heat flux contours.}
    \label{fig:chaTKEheatFlux}
\end{figure}

Attention is next directed to quantities not directly included as optimization targets, TKE and heat-flux distributions, shown in Figure~\ref{fig:chaTKEheatFlux}. In Figure~\ref{fig:chaTKEheatFlux} (a), the baseline RANS model underpredicts TKE and thus provides insufficient turbulent diffusion, producing a narrower jet and weaker temperature rise than DNS, while the CFD-driven model restores the correct spread. Figure~\ref{fig:chaTKEheatFlux} (b) shows that the baseline also underestimates the wall heat-flux near the inner hot wall, which is substantially improved by the CFD-driven model, yielding a more accurate thermal response.

\section{Conclusion}
A surrogate-augmented CFD-driven training framework is proposed to substantially reduce the computational cost of model training. Recognizing that the rich information generated during the CFD-driven training process is underutilized in regular training, the surrogate model leverages these data to establish a mapping between the GEP-generated model parameters and their corresponding cost function values. Instead of evaluating every GEP-generated model through CFD simulations, only those predicted by the surrogate model to have minor errors or large uncertainty ranges are re-evaluated using CFD. The surrogate model itself is also evolved during training. The key components required for developing the surrogate-augmented CFD-driven training approach, including input mapping, surrogate model formulation, and candidate selection criteria for re-evaluation, are described in detail. Various hyperparameter combinations are evaluated on two development cases: a square-duct flow (turbulence-model training only) and a vertical natural-convection (VNC) flow (joint turbulence- and heat-flux-model training), yielding an optimal surrogate configuration. Using this optimized setup, the framework is further assessed on two demonstration cases of increasing complexity, namely horizontal mixed convection (HMC) and concentric horizontal annulus (CHA) flows, by training both the turbulence and heat-flux models and comparing the training error evolution against the regular CFD-driven training. It is found that the surrogate-augmented CFD-driven training framework, using a fixed set of hyperparameters, substantially reduces computational cost over the standard CFD-driven training approach while still producing models with significantly improved predictive accuracy. This improvement greatly enhances the practicality and scalability of CFD-driven model development for more complex, industry-relevant flow applications. 

\section{Acknowledgment}
The authors from the University of Melbourne acknowledge the financial support and publication permission provided by Mitsubishi Heavy Industries, Ltd. Computational resources were supplied by the Pawsey Supercomputing Centre, funded by the Australian Government and the Government of Western Australia under the National Computational Merit Allocation Scheme. 

\appendix
\section{Explicit expressions of trained models}
\label{appendix:models}
The CFD-driven trained turbulence model for the square duct:

\begin{equation}
\begin{aligned}
a_{ij} = 2 \rho k \left ( -1.653 + 0.625I_1 + I_2\right ) V_{ij}^{2}.
\end{aligned}
\label{eqn:ductModel}
\end{equation}

The CFD-driven trained turbulence and heat-flux model for the VNC:

\begin{equation}
\begin{aligned}
a_{ij} = 2 \rho k \left ( -0.098-I_1\right ) V_{ij}^{1}, \\
b_{ij} = 2 \rho k \left ( 2+0.847I_1+I_1^2\right ) V_{ij}^{1}, \\
\alpha^{GEP} = 0.945-2.108I_1.
\end{aligned}
\label{eqn:vncModel}
\end{equation}
$a_{ij}$ and $b_{ij}$ are CFD-driven trained components of the turbulence model, where the former is the extra anisotropy stress and the latter is the artificial stress tensor to produce extra turbulence production. For statistically one-dimensional flow, only the first tensor basis is non-zero. Consequently, only the coefficient of $V_{ij}^{'}$ in $a_{ij}$ and $b_{ij}$, as well as the one $\alpha^{GEP}$ in the heat flux model—three expressions in total—need to be trained. For the trained heat flux model $\alpha^{GEP}=0.945-2.108I_1$, which replaces the baseline value $1/Pr_t$, the corresponding trained $Pr_t \approx 1/0.945 = 1.0582$. This value lies within the commonly used SGDH range, $Pr_t = 0.80 \sim 1.10$ (\cite{xu2021data}), indicating consistency with established modeling practices.

The CFD-driven trained turbulence and heat-flux model for the HMC:

\begin{equation}
\begin{aligned}
a_{ij} = 2 \rho k \left ( -0.404-I_1\right ) V_{ij}^{1}, \\
b_{ij} = 2 \rho k \left ( 0.43+I_1-2I_1^2\right ) V_{ij}^{1}, \\
\alpha^{GEP} = 2.0-J_1(J_1-2.0).
\end{aligned}
\label{eqn:vncModel}
\end{equation}

The CFD-driven trained turbulence and heat-flux model for the CHA:

\begin{equation}
\begin{aligned}
a_{ij} = 2\rho k \Big[ 
   & -\big((N_2 - N_3)(N_3 + 0.089) - 0.85\big) \\
   &  \big(I_1 - I_2(0.089N_1 + 2.205) - 2N_2 + 1.089\big) V_{ij}^{1} \\
   & + \big((I_1 - 3.0)(N_1 + 1.0)\big) V_{ij}^{2}
     + \big(N_3(4I_2 - N_1) - 0.43\big) V_{ij}^{3}
   \Big], \\
   \alpha^{GEP} &= -0.009409 I_1(I_1 + J_1) + 0.71 I_1 + 1.3.
\end{aligned}
\label{eqn:vncModel}
\end{equation}
where $N_1=\mathrm{min} \left( \frac{\sqrt{k}y}{50\nu},2\right)$ is the Reynolds number based on the wall distance, $N_2=\frac{\nu_t}{\nu_t + \nu}$, is the ratio of turbulent to total  (turbulent and molecule) viscosity, and $N_3=F_2$ is the switching function in $k-\omega$ SST.

\section*{Declaration of generative AI and AI-assisted technologies in the manuscript preparation process} During the preparation of this work the author(s) used ChatGPT solely for writing refinement. After using this tool/service, the author(s) reviewed and edited the content as needed and take(s) full responsibility for the content of the published article.

\bibliographystyle{elsarticle-num}
\bibliography{references}
\end{document}